%% file: 0_main.tex
\documentclass[10pt,twocolumn,letterpaper]{article}

\usepackage[pagenumbers]{cvpr} 

\input{0_macros}

\definecolor{cvprblue}{rgb}{0.21,0.49,0.74}
\usepackage[pagebackref,breaklinks,colorlinks,allcolors=cvprblue]{hyperref}

\newcommand{\wlink}[1]{\textcolor{magenta}{{#1}}}

\def\paperID{3640} 
\def\confName{CVPR}
\def\confYear{2025}

\newcommand{\MethodName}{AniGS\xspace}
\title{AniGS: Animatable Gaussian Avatar from a Single Image with \\ Inconsistent Gaussian Reconstruction}

\author{Lingteng Qiu$^{1}\footnotemark[1]$ \quad Shenhao Zhu$^{1,3}\footnotemark[1]$ \quad Qi Zuo$^{1}\footnotemark[1]$ \quad Xiaodong Gu$^{1}\footnotemark[1]$ \\ Yuan Dong$^{1}$ \quad Junfei Zhang$^{1}$ \quad Chao Xu$^{1}$ \quad Zhe Li$^{1,4}$ \quad Weihao Yuan$^{1}$ \quad Liefeng Bo$^{1}$ \\ Guanying Chen$^{2}\footnotemark[2]$ \quad Zilong Dong $^{1}\footnotemark[2]$
\vspace{0.3em} \\
{\normalsize $^1$Alibaba Group}
\quad{\normalsize $^2$Sun Yat-sen University} \\ \quad {\normalsize $^3$Nanjing University} \quad{\normalsize $^4$Huazhong University of Science and Technology}
}

\begin{document}

\twocolumn[{
\renewcommand\twocolumn[1][]{#1}
\maketitle
\vspace{-30pt}
\begin{center}
    \captionsetup{type=figure}
    \includegraphics[width=\textwidth]{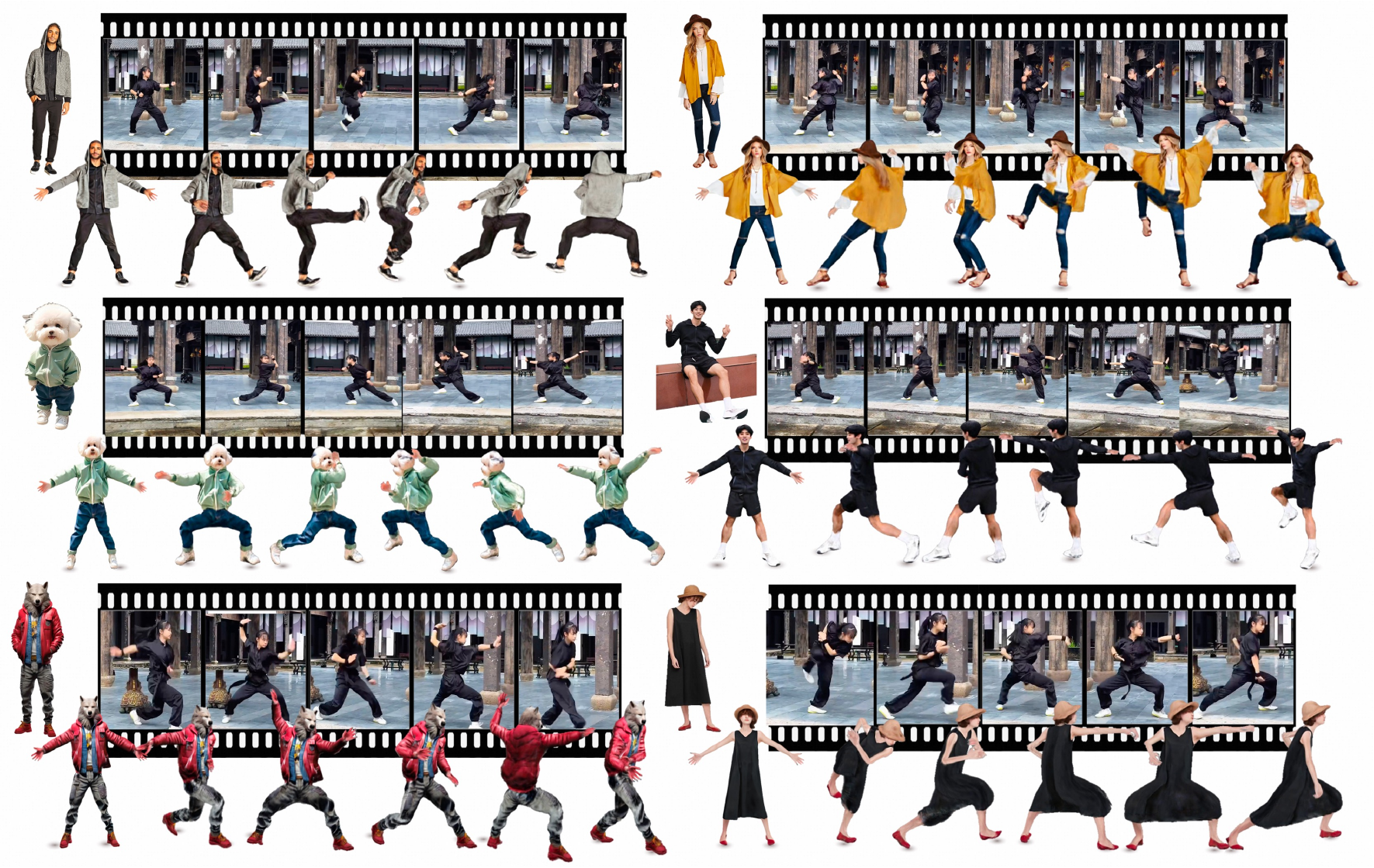}
    \\
    \captionof{figure}{\textbf{3D Avatar Reconstruction and Animation Results of \emph{\MethodName}}. Given a single human image as input, \MethodName is capable of reconstructing a high-fidelity 3D avatar in a canonical pose, which can be used for both photorealistic rendering and real-time animation.}
    \label{fig:teaser}
\end{center}
}]

\footnotetext[1]{Equal contribution.}
\footnotetext[2]{Corresponding author.}

\input{1_abstract}

\input{2_intro}

\input{3_relatedwork}

\input{4_method}

\input{5_experiments}

\input{6_conclusions}

{\small
\bibliographystyle{ieeenat_fullname}
\bibliography{ref}
}
\appendix

\clearpage
\input{7_supp_content}

\end{document}

%% file: 0_macros.tex
\usepackage{xcolor}
\usepackage{enumitem}
\usepackage{xspace}
\usepackage{booktabs}
\usepackage{colortbl}
\usepackage{multirow}
\usepackage{makecell}
\usepackage{lipsum} 
\usepackage{gensymb} 

\newcommand{\Tref}[1]{Table~\ref{#1}}

\newcommand{\fref}[1]{Fig.~\ref{#1}}
\newcommand{\Fref}[1]{Figure~\ref{#1}}

\usepackage[labelsep=period]{caption}
\renewcommand{\paragraph}[1]{\vspace{0.2em}\noindent \textbf{#1 \hspace{0.2em}}}

\definecolor{MyDarkRed}{rgb}{0.76, 0.16, 0.16}
\definecolor{MyDarkBlue}{rgb}{0.16, 0.16, 0.66}

\newcommand{\smplmesh}{\mathcal{M}}

\newcommand{\vaefi}{\mathbb{F}_i}
\newcommand{\vaefv}{\mathbb{F}_v}
\newcommand{\vaefc}{\mathbb{F}_c}
\newcommand{\vaefm}{\mathbb{F}_m}
\newcommand{\smplxnormal}{N_\textrm{smplx}}

\newcommand{\suppl}{supplementary materials\xspace}


%% file: 1_abstract.tex
\begin{abstract}

Generating animatable human avatars from a single image is essential for various digital human modeling applications. Existing 3D reconstruction methods often struggle to capture fine details in animatable models, while generative approaches for controllable animation, though avoiding explicit 3D modeling, suffer from viewpoint inconsistencies in extreme poses and computational inefficiencies.
In this paper, we address these challenges by leveraging the power of generative models to produce detailed multi-view canonical pose images, which help resolve ambiguities in animatable human reconstruction. We then propose a robust method for 3D reconstruction of inconsistent images, enabling real-time rendering during inference.
Specifically, we adapt a transformer-based video generation model to generate multi-view canonical pose images and normal maps, pretraining on a large-scale video dataset to improve generalization. To handle view inconsistencies, we recast the reconstruction problem as a 4D task and introduce an efficient 3D modeling approach using 4D Gaussian Splatting. Experiments demonstrate that our method achieves photorealistic, real-time animation of 3D human avatars from in-the-wild images, showcasing its effectiveness and generalization capability. Our code will be available on \href{https://lingtengqiu.github.io/2024/AniGS/}{\wlink{https://lingtengqiu.github.io/2024/AniGS/}}.

\end{abstract}


%% file: 2_intro.tex

\section{Introduction}
\label{sec:intro}

Generating animatable human avatars has become increasingly important for a wide range of applications, such as virtual reality, gaming, and human-robot interaction. 
However, creating an animatable human avatar with diverse shapes, appearances, and clothing from a single image remains a challenging problem.

Despite significant progress in human reconstruction from a single image~\cite{saito2020pifuhd,xiu2022icon,zhang2024sifusideviewconditionedimplicit,pan2024humansplat}, the reconstructed models are often difficult to animate. This is because the reconstructed human pose is aligned with the input pose, which is usually non-canonical, requiring complex rigging to enable animation. 
For animatable avatar reconstruction, methods based on predicting parametric human models (\eg, SMPL~\cite{loper2015smpl}) with geometry offset refinement often struggle to capture fine details~\cite{alldieck2018videobasedreconstruction3d}. Additionally, methods relying on implicit surface reconstruction face challenges in generalization, primarily due to the lack of large-scale, rigged 3D human datasets for training~\cite{huang2020archanimatablereconstructionclothed,he2022archanimationreadyclothedhuman}.


Recently, controllable human image animation has made significant progress with diffusion models \cite{hu2024animate,zhu2024champ}, which generate animated images directly. These methods achieve realistic results and are easy to animate by input control poses, but suffer from inconsistencies across views due to the lack of a global representation. In addition, these methods also face efficiency issues due to the high computational effort per animation frame.

Motivated by the success of diffusion models in generating multi-view images of objects~\cite{shi2023mvdream,long2024wonder3d}, several methods have explored fine-tuning these models to generate multi-view human images from a single input, followed by 3D reconstruction using multi-view techniques~\cite{he2024magicmangenerativenovelview, liu2024human}. 
Specifically, CharacterGen~\cite{peng2024charactergen} demonstrates the ability to generate \emph{cartoon-style avatars} from a single image by first producing multi-view canonical pose images. To reconstruct the 3D model from these potentially inconsistent multi-view images, a transformer-based 3D reconstruction model is trained. However, training both the multi-view diffusion model and the reconstruction model requires a synthetic 3D \emph{rigged} human dataset to render multi-view canonical images, limiting the generalization ability of these models.

To address these challenges, we adapt a transformer-based video generation model~\cite{hong2022cogvideo} to predict multi-view images and normal maps, by incorporating guidance support from the reference image and human poses. Our generation model can be trained on large-scale, in-the-wild video data, thereby enhancing its generalization.

Despite the high-quality multi-view images generated by our method, inconsistencies still arise, which can affect the 3D reconstruction. By considering these inconsistencies as the dynamic variations within a temporal sequence, we can reformulate the problem of 3D reconstruction from inconsistent images as a 4D reconstruction task. Observing the effectiveness of 4D Gaussian splatting (4DGS) in dynamic scene modeling \cite{Wu_2024_CVPR, yang2024deformable, luiten2023dynamic}, we introduce an efficient 4DGS to fit the multi-view images. After optimization, the high-fidelity 3D avatar model can be obtained as the model in the canonical space of the 4DGS (see~\fref{fig:teaser}).

\vspace{3pt}
In summary, the key contributions of this paper are as follows: 
\begin{itemize}[itemsep=0pt,parsep=0pt,topsep=2bp]
    \item We introduce a method for multi-view canonical pose image generation using a video generation model, trained on unconstrained human pose video data, without the need for synthetic 3D rigged human datasets.
    \item We introduce a new perspective to the problem of 3D animatable avatar reconstruction from inconsistent images, formulating it as a 4D reconstruction task and introducing an efficient 4D Gaussian Splatting model.
    \item Experiments demonstrate that our method generates high-fidelity animatable avatars from a single image, enabling photorealistic and real-time animation during inference.
\end{itemize}





%% file: 3_relatedwork.tex
\section{Related Work}
\label{sec:related_works}

\paragraph{Single-Image Human Reconstruction and Generation} 
Early methods for single-image human reconstruction primarily formulated this problem as geometry offset prediction for mesh-based statistical models for naked~\cite{choutas2022accurate3dbodyshape,kanazawa2018endtoendrecoveryhumanshape,kocabas2020vibevideoinferencehuman,kolotouros2019learningreconstruct3dhuman,saito2021scanimateweaklysupervisedlearning,smith2019facsimilefastaccuratescans,sun2022puttingpeopleplacemonocular} or clothed~\cite{alldieck2019learningreconstructpeopleclothing,alldieck2018detailedhumanavatarsmonocular,alldieck2019tex2shapedetailedhumanbody,lazova2019360degreetexturespeopleclothing,PonsMoll2017ClothCap,xiang2020monoclothcaptemporallycoherentclothing,zhu2019detailedhumanshapeestimation,bhatnagar2019multigarmentnetlearningdress,jiang2020bcnetlearningbodycloth} human body, 
with some approaches extending this to texture prediction~\cite{alldieck2019tex2shapedetailedhumanbody,bhatnagar2019multigarmentnetlearningdress}. 
While a consistent naked topology simplifies animation, it is less effective for modeling diverse clothing styles.
To deal with the cloth-style variation in the wild, numerous notable approaches~\cite{saito2019pifu,saito2020pifuhd,xiu2023econexplicitclothedhumans,xiu2022icon,zheng2020pamirparametricmodelconditionedimplicit,alldieck2022photorealisticmonocular3dreconstruction,corona2023structured3dfeaturesreconstructing,cao2022jiff,zhang2024sifusideviewconditionedimplicit} utilize implicit functions as representations for 3D human models, enabling them to capture complex topologies without suffering from resolution constraints.

Recently, the rise of generative models has blurred the boundaries between reconstruction and generation processes. Some methods~\cite{men2024en3denhancedgenerativemodel,he2024magicmangenerativenovelview,liu2024human} utilize the input image as a conditional element, leveraging generative techniques such as GANs~\cite{men2024en3denhancedgenerativemodel,hong2022eva3dcompositional3dhuman,yang20233dhumangan3dawarehumanimage,dong2023ag3dlearninggenerate3d}, Image Diffusion Models~\cite{chen2024ultramansingleimage3d}, and Video Diffusion Models~\cite{liu2024human,shao2024human4dit} to synthesize 3D human models. 
The concurrent work MagicMan~\cite{he2024magicmangenerativenovelview} has explored the idea of generating multi-view human images and normal maps given the input image.
Unlike these methods, which generate static models, our approach generates multi-view canonical pose images, followed by 3D modeling to reconstruct animatable human models.

\paragraph{Animatable Human Generation} 
Reconstructing animatable avatars remains a challenging problem. Early methods often adopt a parametric-model-based solution~\cite{alldieck2018videobasedreconstruction3d}.
To better model the clothed human body, methods like ARCH~\cite{he2022archanimationreadyclothedhuman,huang2020archanimatablereconstructionclothed} adopts an implicit function representation to represent the geometry of the human body. 
Other works focus on animatable avatar reconstruction from monocular videos~\cite{jiang2022selfrecon,weng2022humannerf, qiu2023recmv, liu2024heromaker} or multi-view videos~\cite{chen2024meshavatar,li2024animatable,peng2021animatable}.
With the advent of text-to-3D techniques~\cite{poole2022dreamfusion}, several methods have explored generating 3D avatars from text prompts~\cite{huang2024dreamwaltz,kolotouros2024dreamhuman,cao2024dreamavatar}.

Recently, CharacterGen~\cite{peng2024charactergen} achieves cartoon-style avatar generation from a single image by generating multi-view images with a diffusion model, followed by transformer-based shape reconstruction. In contrast, our work focuses on in-the-wild animatable human avatars, using a video generation model pretrained on large-scale in-the-wild videos to improve generalization. Additionally, we propose a robust 3D reconstruction method based on 4DGS, avoiding the need for training a transformer reconstruction model.

\begin{figure*}[tb] \centering

    \includegraphics[width=\textwidth]{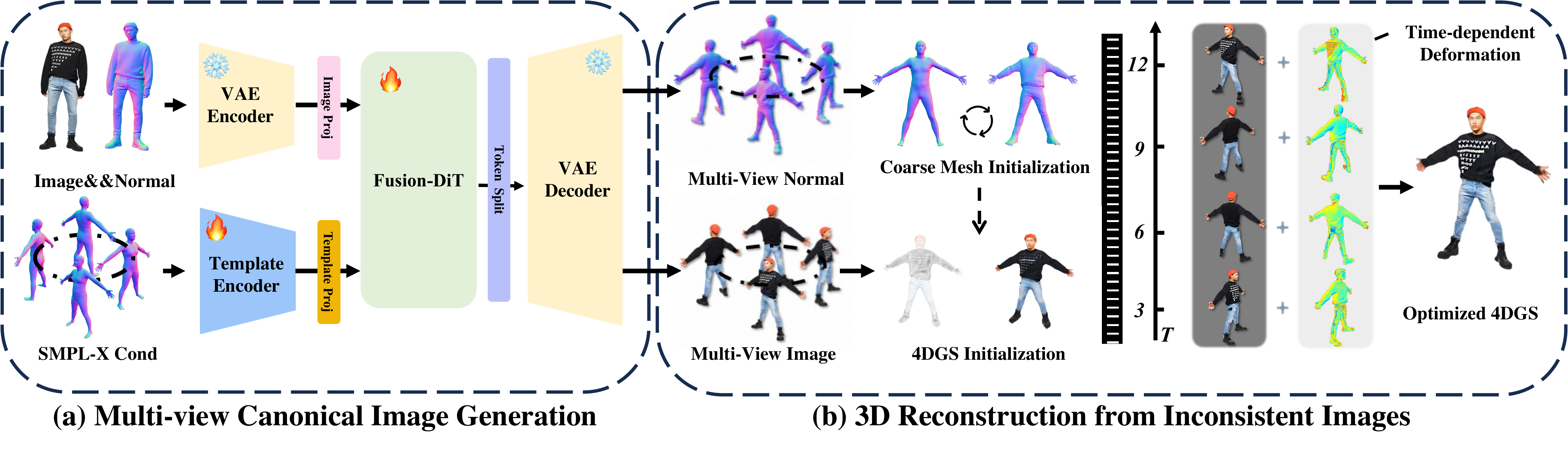}
    \caption{\textbf{Overview of the proposed \MethodName}. 
    In the first stage, a reference image-guided video generation model is employed to produce high-quality multi-view canonical human images along with their corresponding normals, based on the input image. In the second stage, a robust 3D model reconstruction method is applied, using 4D Gaussian Splatting (4DGS) optimization to handle subtle appearance variations across the generated views.
    } 
    \label{fig: overview}
\end{figure*}

\paragraph{Controllable Human Image Animation} 
Original 2D diffusion models~\cite{rombach2022high} primarily focus on generating single-view images and do not support human animation. 
Animate Anyone~\cite{hu2024animate} introduces a reference net into the diffusion models to preserve the identity of the input image and incorporates a lightweight, pose-guided network for guidance.
Champ~\cite{zhu2024champ}, MIMO~\cite{men2024mimo}, and Human4DiT~\cite{shao2024human4dit} employ 3D-level shape guidance~\cite{smplx:2019,loper2015smpl} rather than sparse 2D keypoints, enabling more accurate, controllable image generation under human attributes conditions. 

Although they can produce vivid human-centric animation videos, body distortion, and identity mutation often occur when the animated human turns back at wide angles. 
Additionally, these methods require several minutes to generate a video sequence given the human poses \cite{zhu2024champ,shao2024human4dit}, limiting their practicality in interactive applications.
In contrast, by leveraging an explicit high-fidelity 3D Gaussian model, our methods achieves real-time photorealistic rendering at inference time.

%% file: 4_method.tex
\section{Preliminary}
\paragraph{Human Parametric Model} 
The SMPL~\cite{loper2015smpl} and SMPL-X~\cite{smplx:2019} parametric models are widely used for human body representation. These models utilize skinning techniques and blend shapes derived from a dataset of thousands of 3D body scans. 
Specifically, SMPL-X employs shape parameters \(\boldsymbol{\beta} \in \mathbb{R}^{20}\) and pose parameters \(\boldsymbol{\theta} \in \mathbb{R}^{55 \times 3}\) to represent body mesh deformations.

\paragraph{Diffusion Transformer Model for Video Generation} 
Diffusion Probabilistic Models~\cite{ho2020denoising, song2020denoising} use a forward Markov chain to gradually transform a sample \(x_0\) drawn from the data distribution \(p(x)\) into a noisy equivalent \(q(x)\):
\begin{equation}
q(x_t) = \sqrt{\alpha_t} \cdot {x_0} + \sqrt{1-\alpha_t} \cdot \epsilon, t\in (0, T),
\label{eq:q_sampling}
\end{equation} 
where ${\mathbf{\epsilon}} \sim \mathcal{N}(0, I)$ denotes Gaussian noise, $T$ is the final time step, $t$ is the current time step, and $\alpha_t$ is the noisy schedule parameter. 

The CogVideo model~\cite{hong2022cogvideo, yang2024cogvideox} is an open-source Text-to-Video~(T2V) diffusion model that employs a Transformer architecture~\cite{Peebles2022DiT}, referred to as \(\mu_\theta\), to model the reverse diffusion process. The transformer model processes the current noisy sample \(x_t\), the associated time step \(t\), and optional conditioning inputs \(c\) to estimate the noise \(\epsilon\). 
The loss function for training the denoising model is:
\begin{equation}
\mathcal{L}_{\theta} = \mathbb{E}_{x_0, c, t} \left\| x_0 - {\mu_{\theta}}(x_t, c, t) \right\|^2.
\label{eq:denoise}
\end{equation}

\paragraph{3D Gaussian Splatting} 
3D Gaussian Splatting~\cite{kerbl3Dgaussians} represents 3D data using a collection of 3D Gaussians. Each Gaussian is defined by a center \(\mathbf{x} \in \mathbb{R}^3\), a scaling factor \(\mathbf{s} \in \mathbb{R}^3\), and a rotation quaternion \(\mathbf{q} \in \mathbb{R}^4\). Additionally, it includes an opacity value \(\alpha \in \mathbb{R}\) and a color feature \(\mathbf{c} \in \mathbb{R}^C\) for rendering purposes, with spherical harmonics capturing view-dependent effects. Rendering these Gaussians involves projecting them onto the image plane as 2D Gaussians and applying alpha compositing for each pixel in a front-to-back order. Recent methods~\cite{Wu_2024_CVPR,yang2024deformable,luiten2023dynamic} extend 3D Gaussian Splatting to capture dynamic 4D scenes by incorporating a temporal embedding.


\section{Method}
\label{sec:method}

\subsection{Overview}
\label{sub:Overview}

Given an input human image $I \in \mathbb{R}^{H \times W \times 3}$, our goal is to create an animatable 3D avatar represented by a 3DGS model. This avatar can be animated by applying new human pose conditions during inference. To facilitate animation, it is crucial to reconstruct the 3D avatar in a canonical pose, simplifying the rigging process. The 3D avatar in this canonical pose can be rigged using the aligned SMPL-X model or other off-the-shelf rigging methods \cite{accurig}. As shown in \fref{fig: overview}, our framework consists of two main stages.

In the first stage, we employ a reference image-guided video generation model to produce high-quality multi-view canonical human images and their corresponding normals from the input image. In the second stage, we reconstruct the 3D model using these generated images. However, despite the high quality of the generated images, multi-view inconsistencies still arise due to the nature of the diffusion-based video generation model. Consequently, directly applying traditional multi-view reconstruction methods to these images often results in the loss of detail and the introduction of artifacts.

To tackle the issue of view inconsistencies, we formulate this problem as a 4D reconstruction task and introduce an efficient 4D Gaussian Splatting model to account for the appearance variations at different timesteps (\ie, viewpoints).
After optimization, a high-fidelity 3D model can be obtained as the model in the canonical space of the 4DGS.

\subsection{Multi-view Canonical Image Generation}
\label{method: mv generation}

Given a reference human image in an \emph{arbitrary pose}, our goal in the first stage is to generate multi-view RGB images of the same subject in a \emph{canonical pose}. 
Motivated by recent successes in controllable image generation through video models~\cite{yang2024cogvideox, hu2024animate}, we adapt a diffusion transformer-based video generation model to reposition the human subject to a canonical pose and generate multi-view images.

Specifically, the video generation model takes as input the reference image and SMPL-X pose conditions to produce multi-view images. Here, the rotation of the camera in relation to the subject is treated as equivalent to subject rotation.

\paragraph{Reference-guided Canonical Video Generation}
We extend the CogVideo model~\cite{hong2022cogvideo} to achieve reference image-guided and SMPL-X pose-guided video generation. 
This model includes a transformer-based denoiser and a Variational Autoencoder (VAE) that maps input videos or images into a high-dimensional latent space.

The reference image \(I\) is first encoded into VAE features \(\vaefi \in \mathbb{R}^{B \times 1 \times C \times HW}\), while the latent representations for video noise are denoted by \(\vaefv \in \mathbb{R}^{B \times f \times C \times HW}\), where \(B\) represents the batch size, \(f\) the number of output frames, \(C\) the number of latent feature channels, and \(HW\) the total number of input feature tokens. 

To ensure that generated multi-view images retain the identity of the reference image, we fuse the reference image features and latent features during the denoising process. We achieve this by concatenating the reference image features and latent features along the frame channel, yielding \(\vaefc \in \mathbb{R}^{B \times (f+1) \times C \times HW}\) at each DiT block, which allows for feature interaction through self-attention.

To guide the video generation with input human poses, we integrate a lightweight pose guidance network inspired by CHAMP~\cite{zhu2024champ}. This network extracts guidance features from the canonical SMPL-X normal, \(\smplxnormal\), which are added to the corresponding noisy latents to direct the denoising process.

\paragraph{Joint Multi-view RGB and Normal Generation}
To enhance multi-view reconstruction with normal supervision~\cite{long2024wonder3d, ye2024stablenormal}, we further extend the video generation framework to simultaneously produce multi-view RGB images and normal maps, conditioned on a reference image and its corresponding normal map, as predicted by an existing method~\cite{ye2024stablenormal}.

We employ the CogVideo-2B architecture~\cite{yang2024cogvideox} as our base model, which comprises 30 DiT blocks. We modify the first three DiT blocks into two branches, one for RGB and the other for normal inputs. Similarly, we modify the last three DiT blocks into two branches that simultaneously output multi-view RGB images and normal maps.

To effectively integrate image and normal features, we first share the weights of the middle 24 DiT blocks for both RGB and normal feature processing. Secondly, we insert a multi-modal attention module between every three shared DiT blocks and the head of the middle DiT blocks. For this multi-modal attention block, we concatenate the RGB and normal features at the token level, resulting in \(\vaefm \in \mathbb{R}^{B \times f \times C \times 2HW}\).

\paragraph{Training Strategy}
To enhance the generalization capabilities of our model in the face of limited large-scale synthetic datasets, we first pre-train the video generation model on a large-scale, in-the-wild dataset consisting of 100,000 single-human animation videos. As ground-truth normal maps are unavailable for in-the-wild data, we use Sapiens~\cite{khirodkar2024sapiens} and Multi-HMR~\cite{multi-hmr2024} to generate pseudo labels for both normal maps and SMPL-X human poses. During training, we randomly sample frames from these videos to predict short video segments.

Following pre-training, we generate synthetic 3D data assets to obtain self-rotated RGB images along with corresponding ground-truth normal maps. To maintain the model’s generalizability, we use a training strategy that allocates 10\% probability to in-the-wild data and 90\% to synthetic data.

\subsection{3D Reconstruction from Inconsistent Images}
\label{method: Repurpose 4DGS to mv recon}
\begin{figure}[tb] \centering
    \includegraphics[width=0.48\textwidth]{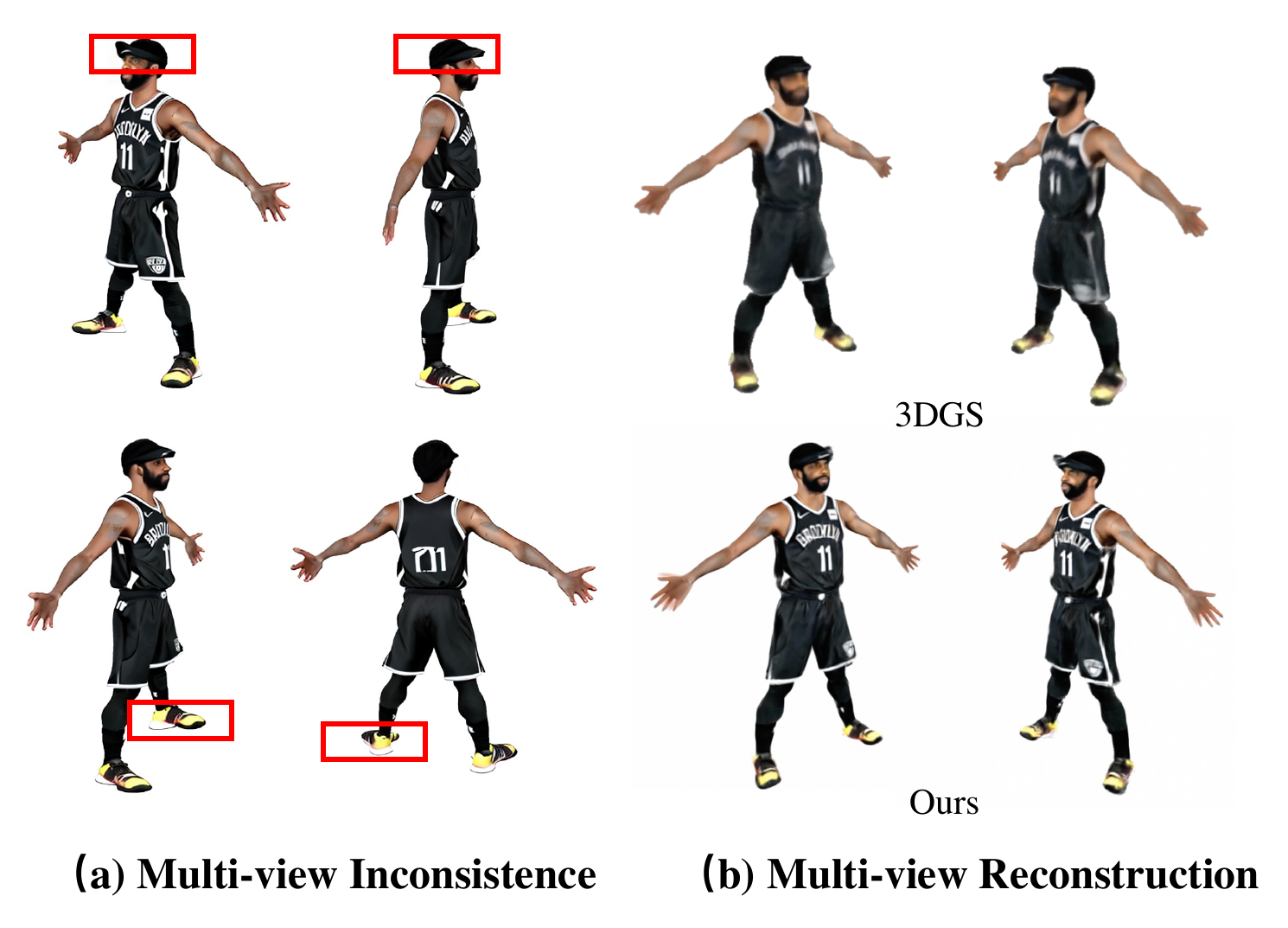}
    \caption{Inconsistencies caused by subtle variations in the generated multi-view images, which will degrade the 3D reconstruction quality. The \fcolorbox{red}{white}{red boxes} highlights the inconsistent areas.} 
    \label{fig: inconsistent multi-view generaiton}
\end{figure}

Once multi-view images are generated by the diffusion model, we can reconstruct a 3D Gaussian human model in canonical space. 
However, due to subtle appearance variations across views in the generated images, directly applying 3D Gaussian Splatting (3DGS) optimization degrades the quality of the reconstructed avatar (see \fref{fig: inconsistent multi-view generaiton}).


\paragraph{Problem Formulation} 
To address the challenges of 3D reconstruction from inconsistent views, it is necessary to handle the shape and appearance variations in each view. 
Viewing these inconsistencies as analogous to dynamic variations within a temporal sequence, we can reformulate the problem of 3D reconstruction from inconsistent images as a 4D reconstruction task. 
Inspired by the recent success of 4D Gaussian Splatting in dynamic scene modeling~\cite{Wu_2024_CVPR, yang2024deformable, luiten2023dynamic}, we adopt a 4DGS approach to achieve efficient optimization and rendering. 

The 4DGS framework is composed of a canonical space and a per-frame deformation module. The canonical space represents a static 3D model (\eg, defined as the first frame), while the per-frame deformation module estimates shape and color variations of each 3D Gaussian, conditioned on the frame index, to fit the video sequence.

Our goal is to optimize both the canonical space and the deformation module based on the generated inconsistent images. 
Once optimized, this process yields a multi-view consistent Gaussian avatar model representing the shape in canonical space.

\paragraph{4D Gaussian Splatting Model}
Following existing dynamic Gaussian splatting methods~\cite{Wu_2024_CVPR}, the deformation of 3D Gaussians is modeled by a deformation field network. 
We employ an efficient spatial-temporal encoder architecture consisting of a multi-resolution HexPlane and a compact MLP~\cite{Cao2023HexPlane, TiNeuVox, shao2023tensor4d}. This structure encodes both the temporal and spatial features information of 3D Gaussians across six 2D voxel planes, incorporating temporal effects. 

Once the 3D Gaussians features are encoded, separate MLPs are employed to compute the deformation for position $\Delta \mathbf{x}$, rotation $\Delta \mathbf{r}$, and scaling $\Delta \mathbf{s}$.
Then, the deformed 3D Gaussian $(\mathbf{x}^\prime,\mathbf{r}^\prime,\mathbf{s}^\prime)$ can be expressed as:
\begin{align}
    & (\mathbf{x}^\prime, \mathbf{r}^\prime, \mathbf{s}^\prime) = (\mathbf{x} + \Delta \mathbf{x}, \mathbf{r} + \Delta \mathbf{r}, \mathbf{s} + \Delta \mathbf{s}) .
\end{align}

\paragraph{Shape Regularization}
In multi-view generation, both RGB images and normal maps are produced, allowing us to regularize the 4DGS optimization process using surface normal regularization. Specifically, we apply an \(L_1\) loss on the rendered normals, \(\mathcal{L}_\textrm{normal}\), to guide the optimization.

To mitigate the occurrence of spikes artifacts during avatar animation, which is often due to excessively large or elongated 3D Gaussian ellipsoids, we incorporate an anisotropy regularizer, $\mathcal{L}_\textrm{ar}$, to constrain the shape of the 3D Gaussians, following~\cite{physgaussian}.

\paragraph{4DGS Optimization} 
In line with existing reconstruction methods~\cite{kerbl3Dgaussians, kplanes_2023, dnerf}, we employ L1 loss for both color $\mathcal{L}_\text{color}$ and mask supervision $\mathcal{L}_\text{mask}$.  
We also include a grid-based total variation loss, \(\mathcal{L}_{tv}\), as proposed in~\cite{Wu_2024_CVPR}, to promote smooth deformation across views. For the human mask labels, we obtain pseudo ground-truth masks using SAM2~\cite{ravi2024sam2}. 
The total reconstruction loss is formulated as:
\begin{equation}
\begin{aligned}
    \mathcal{L}_r &= \mathcal{L}_\text{color} + \lambda_{m} \mathcal{L}_\text{mask} + \lambda_{n} \mathcal{L}_\text{normal} + \lambda_{tv} \mathcal{L}_{tv} + \lambda_{ar} \cdot \mathcal{L}_{ar},
\end{aligned}
\end{equation}
where the weights for different loss terms are set as \(\lambda_{m} = 0.1\), \(\lambda_{n} = 0.05\), \(\lambda_{ar} = 0.001\), and \(\lambda_{tv} = 1\).

\paragraph{Point Cloud Initialization of 4DGS}
Point cloud initialization is critical for 3DGS optimization. We begin by generating a coarse mesh from the multi-view images and sample points on its surface to initialize the 3DGS points. 

Specifically, we deform the predicted SMPL-X mesh to fit the generated multi-view RGB masks and normal maps. We use the Nvidiffrast rasterizer~\cite{Laine2020diffrast} to render both the foreground and normal map of the deformed mesh. To reduce artifacts caused by inconsistencies in the multi-view images, we apply a Laplacian loss and an edge loss to regularize the mesh deformation. Further details are provided in the \suppl.

\subsection{Animation} 
The reconstructed avatar is represented in a canonical space, aligned spatially with the canonical pose of the human body parametric model, SMPL-X. This alignment enables us to use SMPL-X driving parameters to animate the reconstructed avatar, making it fully animatable in 3D. To better handle cloth with large deformation, we apply a diffusion-based skinning method~\cite{lin2022fite, qiu2023recmv}, which transfers SMPL-X skinning weights throughout the entire human canonical space. During animation, skinning weights are obtained by querying the weights in space via bilinear interpolation. Additional details are provided in the \suppl.

\begin{figure*}[htb] \centering
    \includegraphics[width=0.9\textwidth]{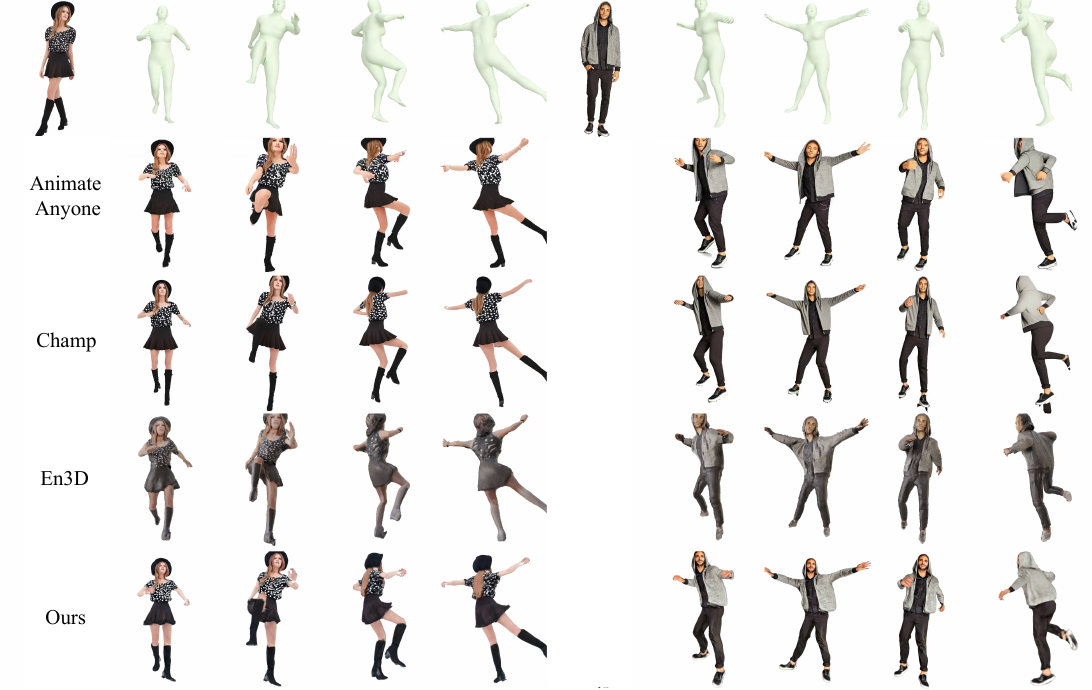}
    \caption{Visual comparison of animation results for the reconstructed 3D avatars. Best viewed with zoom-in.} 
    \label{fig: qualitative results about Animation results.}
\end{figure*}

%% file: 5_experiments.tex
\section{Experiments}
\label{sec:Experiments}
In this section, we thoroughly evaluate the effectiveness of our proposed methods by conducting a comprehensive comparison with state-of-the-art approaches.

\paragraph{Training Datasets} 
For training the multi-view generation model, we first conduct a pretraining phase using a large dataset of dynamic human videos. Specifically, we collect approximately 200,000 dynamic human videos from various online sources. From this, we manually select single-person videos to create our in-the-wild training dataset, which consists of around 100,000 video samples.
During the fine-tuning phase, we leverage a combination of public synthetic 3D datasets to render multi-view images.
These datasets include 2K2K~\cite{han2023highfidelity3dhumandigitization}, Thuman2.0, Thuman2.1~\cite{tao2021function4d}, and CustomHumans~\cite{ho2023custom}, along with commercial datasets such as Thwindom and RenderPeople. 
Note that no rigged human models are used for training. In total, we utilize 6,124 synthetic human scans.
%


\paragraph{Inference} 
Our model can generate an animatable 3D avatar in a canonical pose within a few minutes using a single RTX-3090 GPU. Specifically, it takes approximately 5 minutes to generate 30 frames of multi-view RGB and normal images. The 4DGS optimization process takes around 5 minutes. Once the multi-view reconstruction is complete, we set the time parameter
\( t = 0 \) to obtain the final Gaussian point clouds. After optimization, the avatar can be animated and rendered in real time.


\begin{figure*}[tb] \centering
    \includegraphics[width=\textwidth]{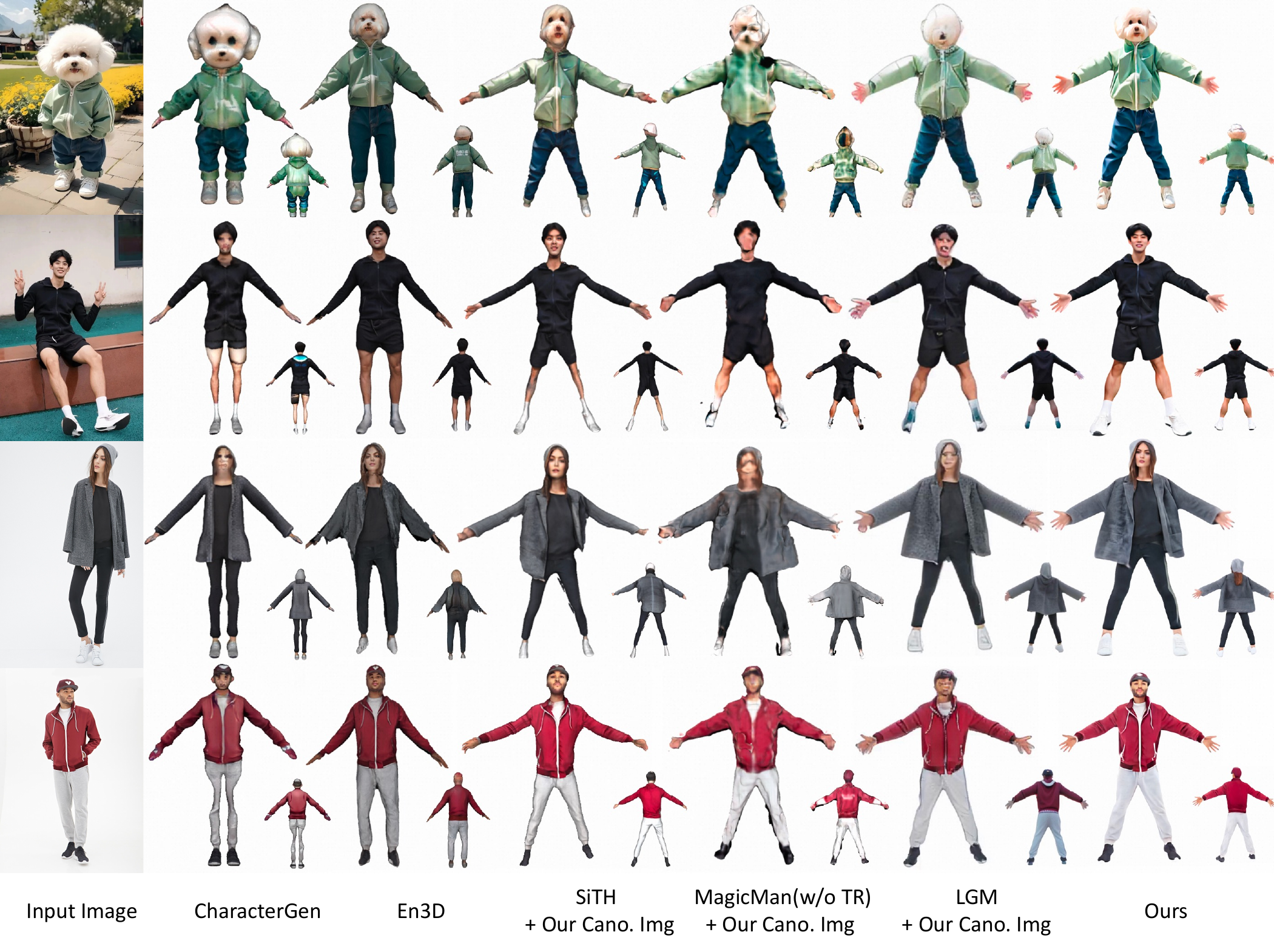}
    \\
    \vspace{-0.5em}
    \caption{Visual comparison on canonical pose 3D avatar reconstruction from the single-view image. Since SiTH, MagicMan, and LGM cannot reconstruct canonical pose shapes from the input, we take our generated front-view canonical pose image as input to these methods.} 
    \label{fig:res_3d_recon}
\end{figure*}

\subsection{Comparison with Existing Methods}

\paragraph{Evaluation Dataset} We choose 50 rigged human avatars from Human4DiT~\cite{shao2024human4dit} to evaluate our performance on multi-view canonical-pose image generation and 3D reconstruction metrics. For animation metrics, we use Blender software to obtain ground-truth video sequences and export the motion sequence to drive the created human models. We then compute photometric metrics in the foreground area to evaluate our performance on animation.

\paragraph{Baselines} Our pipeline includes canonical multi-view generation, multi-view reconstruction, and human animation. For the multi-view generation task, we choose the state-of-the-art MagicMan~\cite{he2024magicmangenerativenovelview} and CHAMP~\cite{zhu2024champ} as baselines. Notably, since CHAMP does not specifically train on self-rotated synthetic human datasets, we fine-tune this approach on our rendering datasets for a fair comparison.

For multi-view canonical human reconstruction tasks, we compare with the animatable human generation methods~\cite{men2024en3denhancedgenerativemodel}, CharacterGen~\cite{peng2024charactergen}. Additionally, we also conducted comparison experiments with static clothed human reconstruction methods, including SiTH~\cite{ho2024sith}, MagicMan~\cite{he2024magicmangenerativenovelview}, and LGM~\cite{tang2025lgm} using our generated canonical pose image input.
For human animation, we conduct comparison experiments with En3D and CharacterGen on the synthetic dataset.

\begin{table}[tb]\centering
    \caption{Quantitative results on multi-view canonical image generation. $^+$ denotes using our generated canonical pose image as input, and * indicates that we fine-tune this approach on the multi-view synthetic dataset. Normal Loss is the cosine similarity loss between the predicted and ground-truth normal. 
    }
    \label{tab:res_mv_generation}
    \resizebox{0.48\textwidth}{!}{
    \large
    \begin{tabular}{*{10}{c}}
        \toprule
       Methods & PSNR $\uparrow$ &  SSIM $\uparrow$ & LPIPS $\downarrow$ & Normal Loss $\downarrow$ \\
        \midrule
        MagicMan$^+$ ~\cite{zhu2024champ} & \textbf{23.775} &  \textbf{0.911} & 0.1077 & 0.134 &  \\
        \midrule
        CHAMP*~\cite{zhu2024champ} & 20.025 & 0.894 & 0.1458 & - &  \\
        Ours & 23.125 & 0.907  & \textbf{0.1023} & \textbf{0.101} \\
        \bottomrule
    \end{tabular}
    }
\end{table}

\begin{table}[tb]\centering
    \caption{Quantitative comparison of 3D modeling on Human4DiT Synthetic Datasets. 
    }
    \label{tab:res_reconstruction}
    \resizebox{0.48\textwidth}{!}{
    \large
    \begin{tabular}{*{10}{c}}
        \toprule
       Methods & LPIPS $\downarrow$ &  CLIP score $\uparrow$ & FID $\downarrow$ & User $\uparrow$\\
        \midrule
        SiTH~\cite{zhang2024sifusideviewconditionedimplicit} & 0.1607 & 86.854 & 86.895 &  2.134 \\
        LGM~\cite{tang2025lgm} & 0.1567 & 86.686 & 82.121 &  2.617\\
        MagicMan~\cite{he2024magicmangenerativenovelview}  & 0.1479 & 82.693 & 144.642 & 2.095 \\
        \midrule
        CharacterGen~\cite{peng2024charactergen} & 0.1638 & 87.154 &  88.751 & 2.481\\
        En3D~\cite{men2024en3denhancedgenerativemodel} & 0.1576 & 82.975 &  131.32 & 2.549\\
        Ours & \textbf{0.1085} & \textbf{90.370} &  \textbf{77.879} & \textbf{4.199}\\
        \bottomrule
    \end{tabular}
    }
\end{table}

\paragraph{Evaluation on Multi-view Canonical Generation} 
As reported in \Tref{tab:res_mv_generation}, our method outperforms CHAMP in multi-view image generation, clearly demonstrating its effectiveness.
It is important to note that MagicMan~\cite{he2024magicmangenerativenovelview} is specifically designed for multi-view generation of static humans and cannot generate canonical pose images. Therefore, we use our generated canonical pose images as input for MagicMan. While MagicMan achieves better PSNR metrics, it is not suitable for animatable avatar generation. Additionally, our method achieves lower normal errors in the generated normal maps, further emphasizing its superior performance.

\paragraph{Evaluation on Canonical Shape Reconstruction} 
To assess the quality of canonical shape reconstruction, we render each reconstructed model from 24 different views and compute LPIPS and FID scores. The average CLIP score is computed by comparing the input RGB image with all rendered views. 

Since this evaluation is closely related to generation tasks, we also conduct a user study to assess the similarity between the generated model and the ground truth. We randomly sample 20 cases and carry out an anonymous ranking survey with $40$ participants across all baselines. The final metric represents the average ranking score, with a maximum score of 5 for each model.
As shown in \Tref{tab:res_reconstruction}, our method consistently outperforms the baselines on all the metrics, justifying its design.

As demonstrated in \Fref{fig:res_3d_recon}, our method produces significantly better canonical pose shape reconstructions than previous methods, with higher fidelity to the input image and enhanced sharpness in both texture and body shape details (e.g., clothing, faces, and hands). Note that since the texture refinement code for MagicMan~\cite{he2024magicmangenerativenovelview} has not been released, the 3D reconstruction results for MagicMan do not include texture refinement.

\begin{table}[tb]\centering
    \caption{Quantitative results on human animation.}
    \label{tab:res_animation}
    \resizebox{0.40\textwidth}{!}{
    \large
    \begin{tabular}{*{10}{c}}
        \toprule
       Methods & PSNR $\uparrow$ &  SSIM $\uparrow$ & LPIPS $\downarrow$\\
        \midrule
        CharacterGen ~\cite{peng2024charactergen} & 17.570 & 0.644 & 0.205  \\
        En3D ~\cite{he2024magicmangenerativenovelview} & 19.244 & 0.751 & 0.174 &    \\
        Ours & \textbf{21.475} & \textbf{0.857} & \textbf{0.137} & \\
        \bottomrule
    \end{tabular}
    }
\end{table}

\paragraph{Evaluation on Human Animation} 
We compare the animation sequences of different methods with the rendered ground-truth sequences.
As is demonstrated in \Tref{tab:res_animation}, our method outperforms the baseline methods in terms of rendering quality in the animation sequences. 
Compared to the second-best method En3D, our method achieves improvements of 2.231, 0.106, and 0.037 in PSNR, SSIM, and LIPIS, respectively. 
As is visualized in \fref{fig: qualitative results about Animation results.}, our method produces accurate and photorealistic animation results than the baseline methods.
More results are included in the \suppl.




%
%
%
%
%

\subsection{Ablation Study}

\begin{figure}[ht] \centering
    \includegraphics[width=0.48\textwidth]{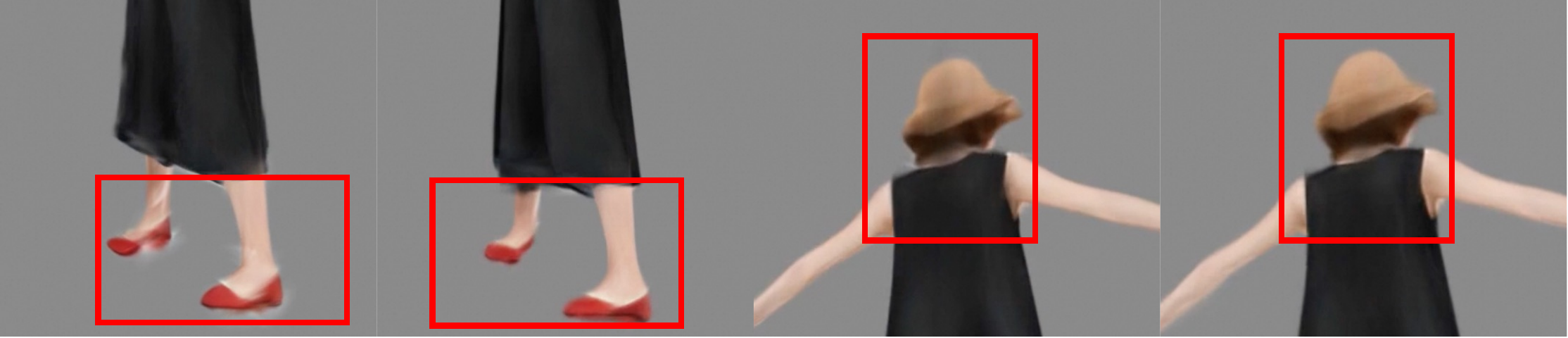}
    \\
    \vspace{-0.4em}
    \makebox[0.11\textwidth]{\footnotesize w/o $\mathcal{L}_\textrm{normal}$ }
    \makebox[0.11\textwidth]{\footnotesize w/ $\mathcal{L}_\textrm{normal}$}
    \makebox[0.11\textwidth]{\footnotesize w/o $\mathcal{L}_\textrm{normal}$} 
    \makebox[0.11\textwidth]{\footnotesize w/ $\mathcal{L}_\textrm{normal}$}
    \\
    \vspace{-0.1em}
    \makebox[0.48\textwidth]{\footnotesize (a) Effects of the normal regularization $\mathcal{L}_\textrm{normal}$ }
    \\
    \vspace{-0.1em}
    \includegraphics[width=0.48\textwidth]{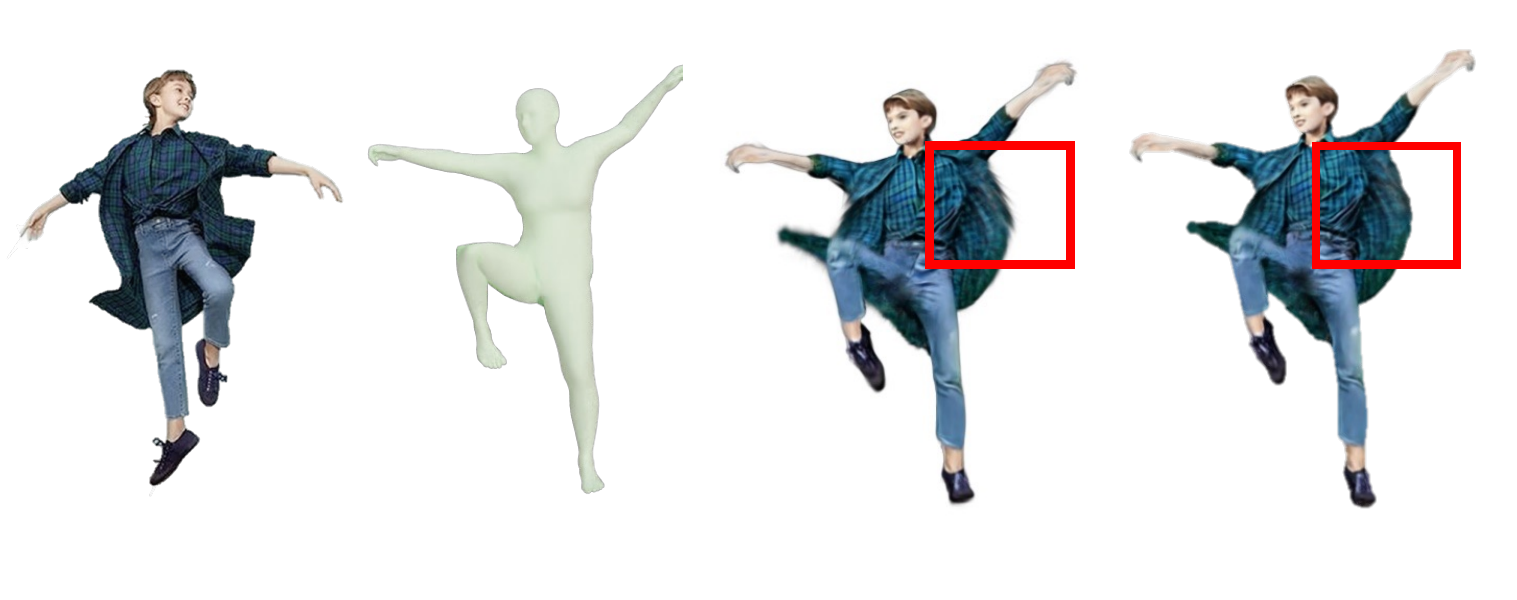} \\
    \vspace{-2em}
    \makebox[0.11\textwidth]{\footnotesize Input }
    \makebox[0.11\textwidth]{\footnotesize Pose }
    \makebox[0.11\textwidth]{\footnotesize w/o $\mathcal{L}_{ar}$} 
    \makebox[0.11\textwidth]{\footnotesize w/ $\mathcal{L}_{ar}$ }
    \\
    \makebox[0.48\textwidth]{\footnotesize (b) Effects of the anisotropic regularization $\mathcal{L}_{ar}$ }
    \\
    \caption{Ablation study for the shape regularization.} 
    \label{fig:shape_reg}
    \vspace{-1em}
\end{figure}

\paragraph{Shape Regularization} 
We conduct an ablation study to evaluate the design of the shape regularization. 
Specifically, \fref{fig:shape_reg}~(a) demonstrates that normal regularization effectively reduces random noise and enhances surface details, while \fref{fig:shape_reg}~(b) shows that anisotropic regularization helps eliminate spikes in novel pose animations.

\begin{figure}[t] \centering
    \includegraphics[width=0.48\textwidth]{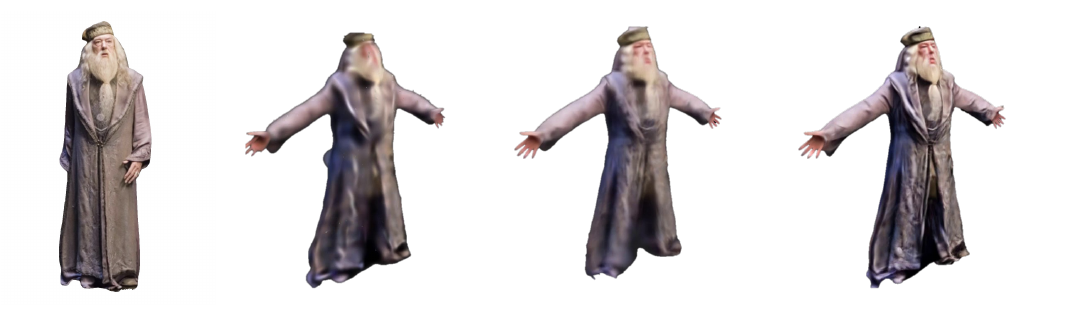}
    \\
    \vspace{-1em} 
    \makebox[0.110\textwidth]{\footnotesize Input Image \hfill}
    \makebox[0.110\textwidth]{\footnotesize Random Init. \hfill}
    \makebox[0.110\textwidth]{\footnotesize SMPL Init.} 
    \makebox[0.110\textwidth]{\footnotesize Our Init.}
    \\
    \caption{Ablation study for shape initialization strategy. } 
    \label{fig:shape_initialization}
    \vspace{-1em}
\end{figure}

\paragraph{Initialization of 4DGS}
\Fref{fig:shape_initialization} compares the results of 4DGS optimization using random points, SMPL mesh, and the proposed coarse mesh as initialization. 

Without any shape prior, the Gaussian model initialized with random points struggles to optimize, resulting in blurry outputs. Initializing with the SMPL mesh provides a better shape prior for body representation but faces difficulties in accurately modeling points that are farther from the body.
In contrast, the coarse mesh initialization provides a good starting point for the optimization.

%% file: 6_conclusions.tex
\section{Conclusion}
\label{sec:Conclusion}
In this work, we present a robust approach to generate animatable human avatars from a single image. 
We introduce a reference image-guided video generation model to produce high-quality multi-view canonical human images and their corresponding normal maps. 
To handle view inconsistencies, we propose a 4D Gaussian Splatting (4DGS)-based method for reconstructing high-fidelity 3D avatars. 
Comprehensive evaluation demonstrates that our method enables photorealistic, real-time animation of 3D human avatars from in-the-wild images.

\paragraph{Limitations and Future Work} 
While our method supports real-time inference, it still requires several minutes to optimize an animatable avatar. In future work, we aim to explore feed-forward 3D reconstruction techniques that are robust to multi-view inconsistencies.

%% file: 7_supp_content.tex
\section{Demo Video}
Please kindly check the \href{https://www.youtube.com/watch?v=e9qgMMvYMr4&t=1s}{Demo Video} for animation results of the reconstructed 3D avatar.

\section{More Details for the Method}
\subsection{Implementation Details}
\paragraph{Multi-view Generation}

For training the multi-view canonical image generation model, we first pre-train our RGB-Normal DiT model on in-the-wild video clips. To supervise the normal map output, we utilize Sapiens~\cite{khirodkar2024sapiens}, an off-the-shelf normal estimation prior, to generate pseudo ground-truth normals from in-the-wild data. The model is trained using the Adam optimizer~\cite{Kingma2014AdamAM} with a learning rate of $2 \times 10^{-4}$ and a batch size of 1. We employ 16 Nvidia A100 80G GPUs for training, with the pre-training process comprising 100,000 optimization iterations. Subsequently, the model is fine-tuned on a synthetic dataset using the same hyperparameters, performing an additional 50,000 iterations of optimization. To preserve the model's generalizability, we adopt a data-mixing strategy during fine-tuning, assigning a 10\% probability to sampling in-the-wild data and a 90\% probability to synthetic data.

\begin{figure*}[htb] \centering
    \includegraphics[width=\textwidth]{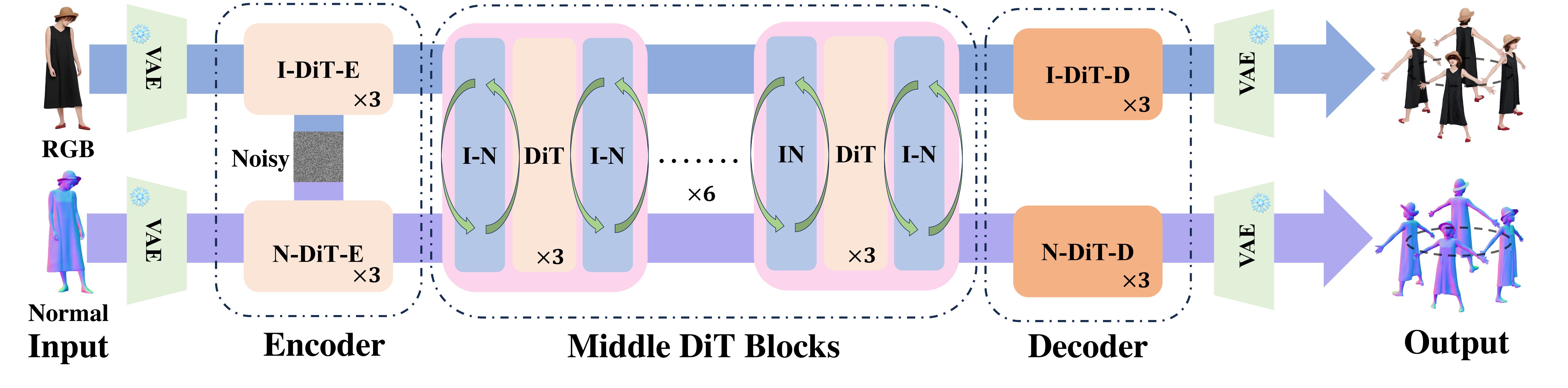}
    \caption{The architecture of the joint RGB-Normal Diffusion Transformer designed for generating multi-view canonical images and normal maps. For simplicity, SPML-X conditioning is omitted from the depiction.} 
    \label{fig:architecture}
\end{figure*}

\paragraph{3D Reconstruction from Inconsistent Images.} In the multi-view reconstruction phase, after obtaining the deformed coarse mesh from the original SMPL-X as the initialization for 4DGS, we first performed 3,000 iterations of optimization the 3DGS parameters. 
Sequentially, we continue to conduct 4,000 iterations of optimization in the temporal dimension to address multi-view inconsistency. 
In the multi-view reconstruction phase, we initialize with a deformed coarse mesh derived from the original SMPL-X model for the 4DGS process. 
The first step is optimizing the 3DGS parameters over 3,000 iterations. Subsequently, we perform 4,000 iterations of optimization considering the temporal dimension to address multi-view inconsistency.

\subsection{RGB-Normal Diffusion Transformer}
\Fref{fig:architecture} illustrates the architecture of our multi-view diffusion transformer model for canonical image and normal map generation. For simplicity, we omit SPML-X conditioning in the figure. Both `I-DiT-E' and `N-DiT-E' denote two independent DiT encoder blocks conditioned on image and normal input, respectively, while `I-DiT-D' and `N-DiT-D' refer to two independent decoders responsible for generating multi-view canonical images and normal maps. Additionally, `I-N' within the intermediate DiT blocks represents a multi-modal attention module that effectively encodes joint image and normal features.

\subsection{Coarse Shape Initialization}%
\label{sub:subsection name}

We optimize the following objective function to obtain the initial coarse mesh $\smplmesh^\prime$ for 3DGS initialization:
\begin{equation}
\begin{aligned}
    \mathcal{L}_{init} &= \lambda_{mask} \cdot \mathcal{L}_{mask} + \lambda_{n} \cdot \mathcal{L}_{normal} \\
    & \quad + \lambda_{lap} \cdot \mathcal{L}_{lap}(\smplmesh^\prime) + \lambda_{edge} \cdot \mathcal{L}_{edge}(\smplmesh^\prime).
\end{aligned}
\end{equation}
where $\lambda_{mask} = 1.0$, $\lambda_{n} = 0.5$, $\lambda_{lap} = 0.1$, and $\lambda_{edge} = 0.05$.

\begin{figure}[tb] \centering
    \includegraphics[width=0.48\textwidth]{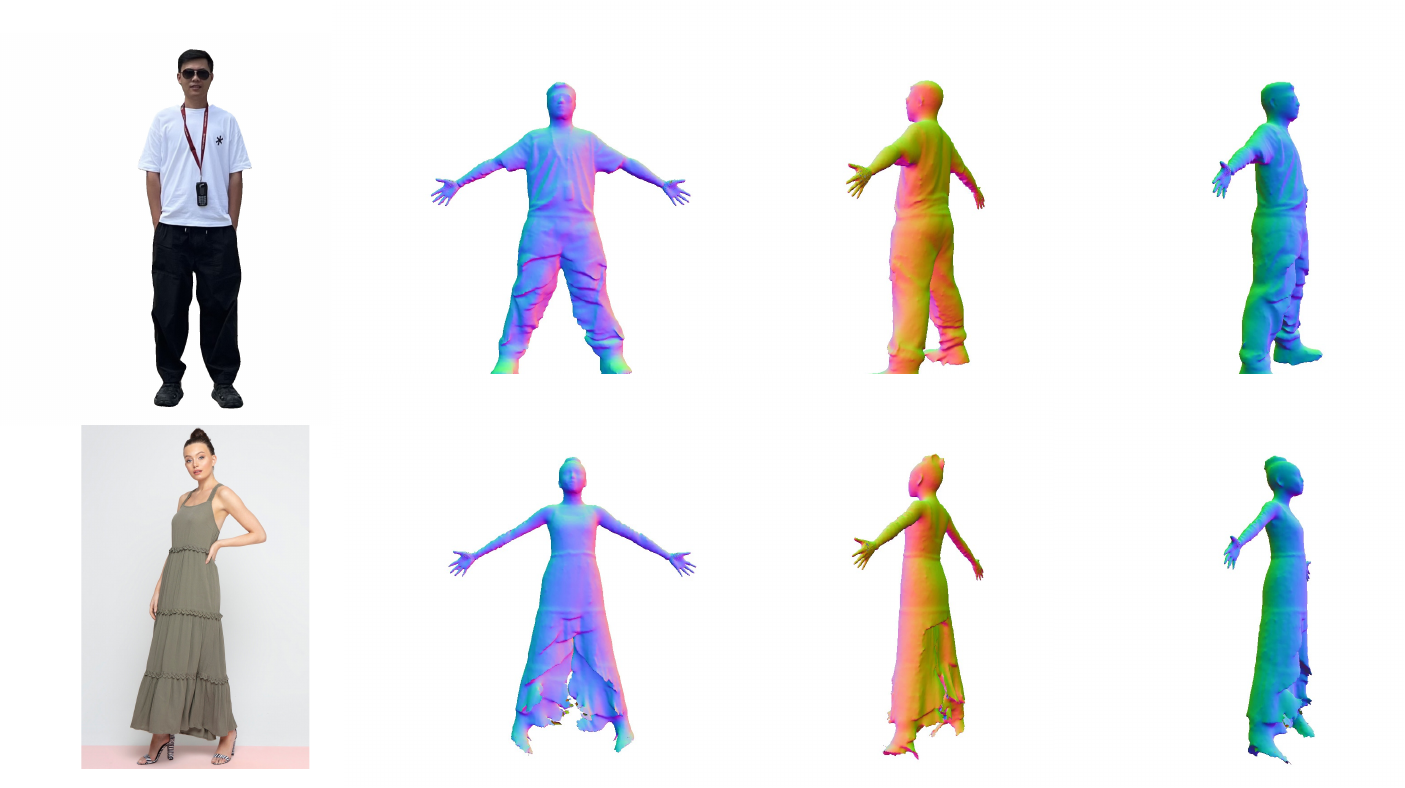}
    \\
        \makebox[0.1\textwidth]{\footnotesize (a) Input}
    \makebox[0.33\textwidth]{\footnotesize (b) The normal map of Coarse Mesh}
    \caption{Sample results for the about coarse mesh reconstruction from multi-view images.} 
    \label{fig:coarse_mesh}
\end{figure}

\Fref{fig:coarse_mesh} demonstrates the coarse mesh results reconstructed from the generated images. As illustrated in the figure, the coarse mesh provides only a rough geometric surface, with several noticeable artifacts remaining on its surface.

\begin{figure}[tb] \centering
    \includegraphics[width=0.38\textwidth]{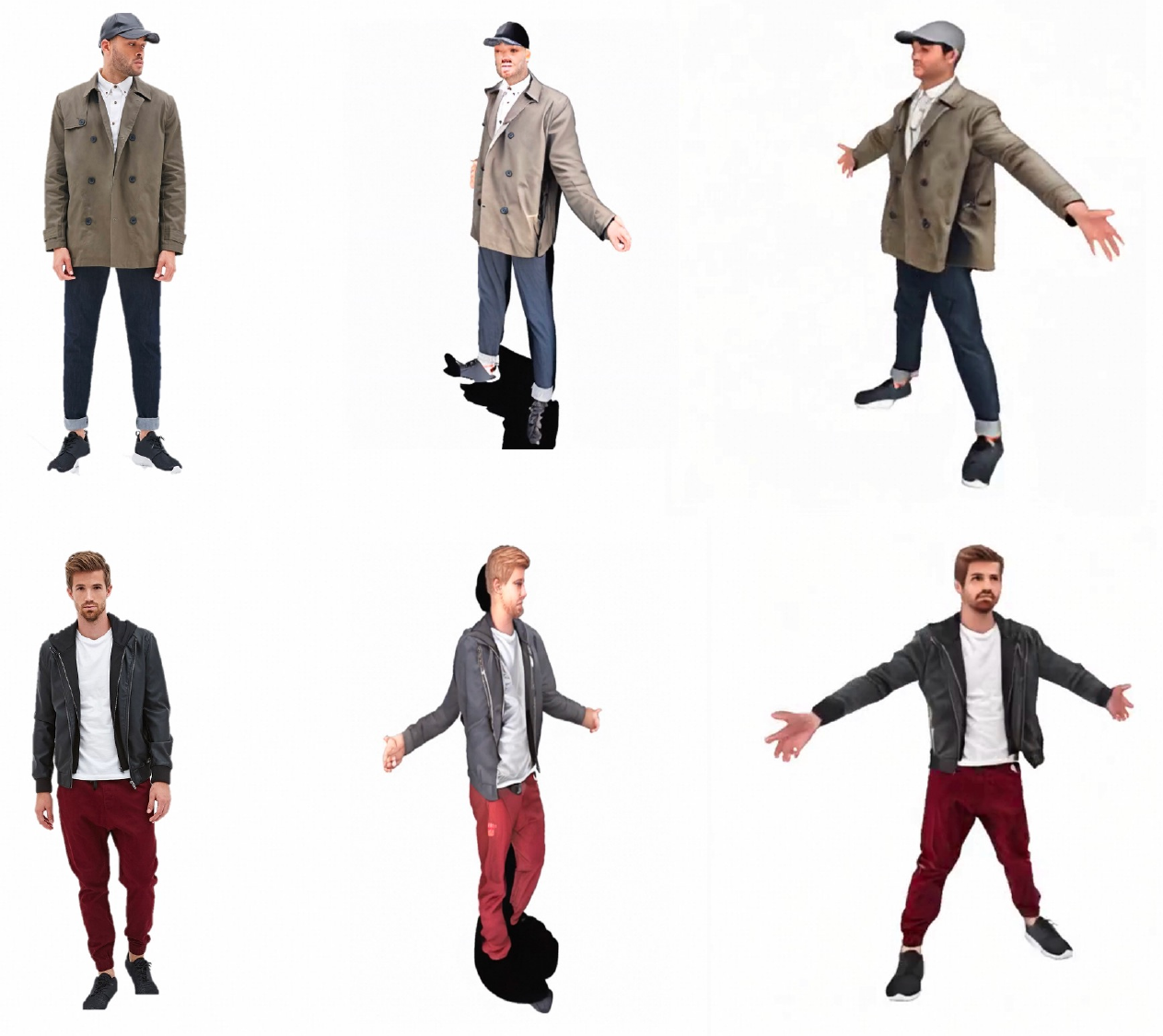}
    \\
   \vspace{-1em}
    \makebox[0.15\textwidth]{\footnotesize Reference }
    \makebox[0.15\textwidth]{\footnotesize (b) w/o pretraining}
    \makebox[0.15\textwidth]{\footnotesize (c) w/ pretraining} 
   \\ 
    \caption{Effectiveness of pre-training on in-the-wild videos. } 
    \label{fig:pretraining}
\end{figure}

\subsection{Skinning-based Animation}

We model large body motions using linear blend skinning (LBS) transformations based on the SMPL-X~\cite{smplx:2019} model. Specifically, given an SMPL body with shape parameter $\beta$ and pose parameter $\theta_i$ in the $i$-th frame, a point $p$ on the body surface in canonical space with skinning weights $w(p)$ can be warped to camera view space via the skinning transformation $W$.

Notably, the skinning weights $w(p)$ are only defined for points on the SMPL-X surface. 
To handle shapes with large deformations (\eg, skirts) and to better facilitate
 the warping of arbitrary points in canonical space to the camera view, we employ the diffused skinning strategy~\cite{lin2022fite} to propagate the skinning weights of the SMPL-X body vertices to the entire canonical space. These weights are stored in a voxel grid of size $256 \times 256 \times 256$. Skinning weights for arbitrary points are then obtained through trilinear interpolation.

%
%
\subsection{More Details for the Synthetic Dataset}

We leverage a combination of public synthetic 3D datasets to render multi-view images for fine-tuning the multi-view canonical image and normal generation model. These datasets include 2K2K~\cite{han2023highfidelity3dhumandigitization}, Thuman2.0, Thuman2.1~\cite{tao2021function4d}, and CustomHumans~\cite{ho2023custom}, along with commercial datasets such as Thwindom and RenderPeople. In total, we utilize 6,124 synthetic human scans.

For the synthetic data, we render each object from 30 different viewpoints by rotating the object. To improve the quality of multi-view reconstruction, images are rendered at varying elevations, which helps to regularize the optimization of the 3D Gaussian Splatting (3DGS) method. Specifically, the elevation range oscillates between $-20^\circ$ and $20^\circ$, following a sine function over a cycle of 30 views.

\section{More Results}

\subsection{Pre-training on In-the-wild Data} 
\Fref{fig:pretraining} underscores the critical role of pre-training on in-the-wild data. 
Models pre-trained on diverse and real-world datasets demonstrate substantially enhanced generalization capabilities compared to models trained without pre-training, verifying the training strategy of our method.

\subsection{Animation Results}

\begin{figure*}[tb] \centering
    \includegraphics[width=\textwidth]{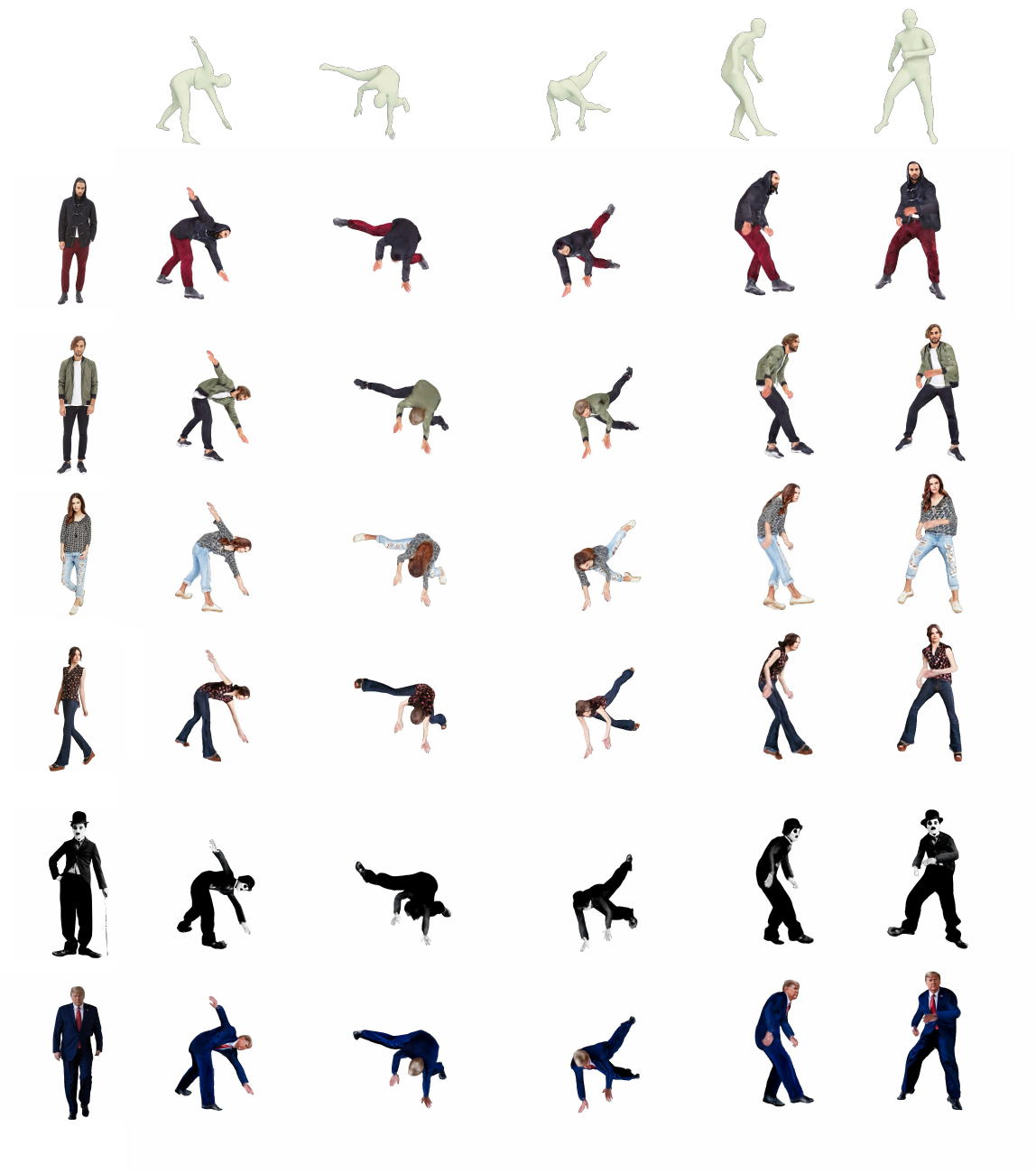}
    \\
    \makebox[0.08\textwidth]{\footnotesize Reference}
    \makebox[0.8\textwidth]{\footnotesize Animation results} 
   \\ 
    \caption{Visual results of human animation results~(Part I) from any input. Best viewed with zoom-in.} 
    \label{fig:animation_v1}
\end{figure*}

\begin{figure*}[tb] \centering
    \includegraphics[width=\textwidth]{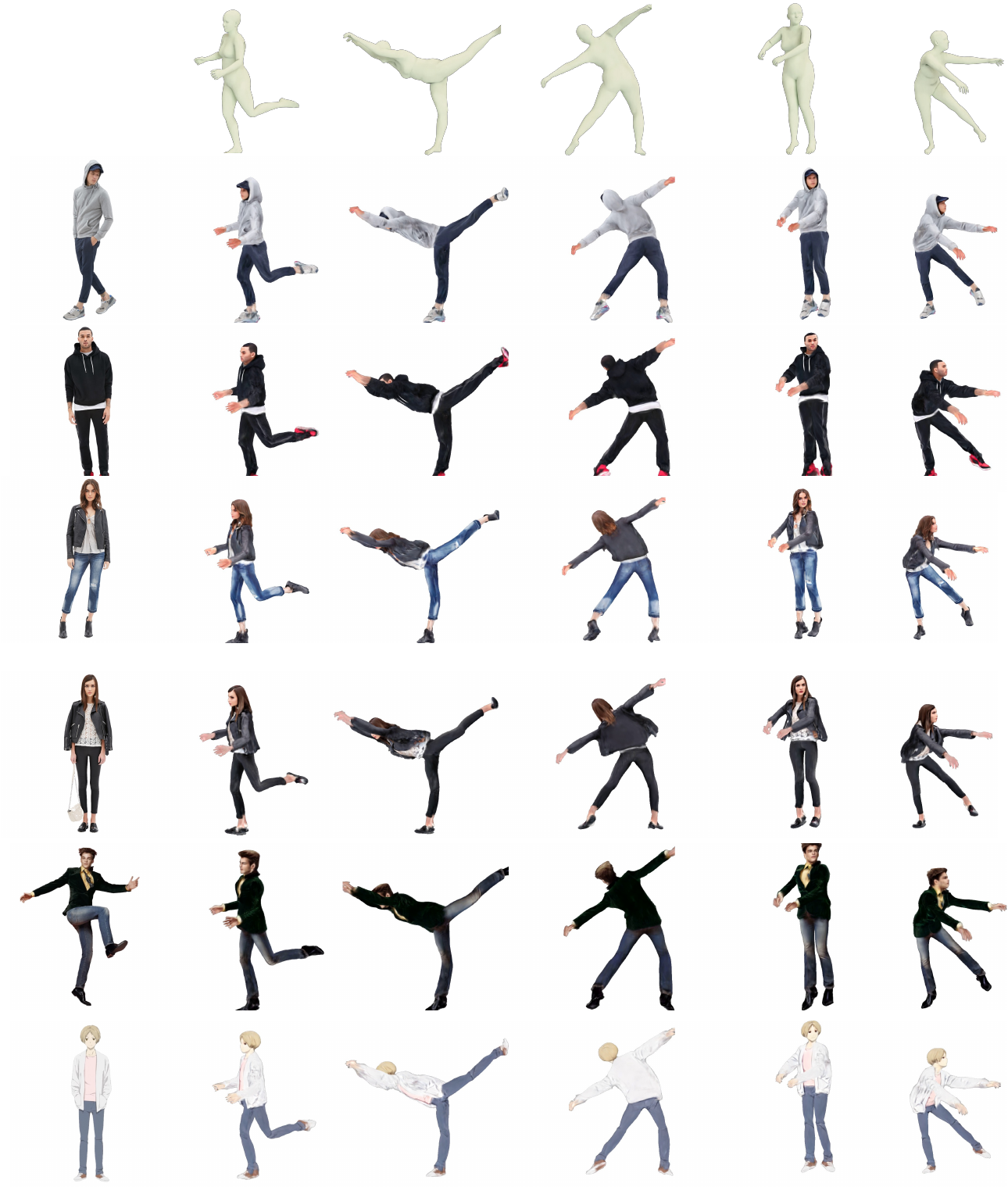}
    \\
    \makebox[0.08\textwidth]{\footnotesize Reference}
    \makebox[0.8\textwidth]{\footnotesize Animation results} 
   \\ 
    \caption{Visual results of human animation results~(Part II) from any input. Best viewed with zoom-in.} 
    \label{fig:animation_v2}
\end{figure*}

\Fref{fig:animation_v1}--\Fref{fig:animation_v2} showcase the animation results of input human images with diverse appearances and a wide range of poses. Our method demonstrates the ability to generate animations that are both robust and photorealistic, preserving fine details of the human appearance while ensuring smooth and natural motion transitions. These results highlight the generalizability and effectiveness of our approach in handling varying levels of complexity in human avatars.

\begin{figure*}[tb] \centering
    \includegraphics[width=\textwidth]{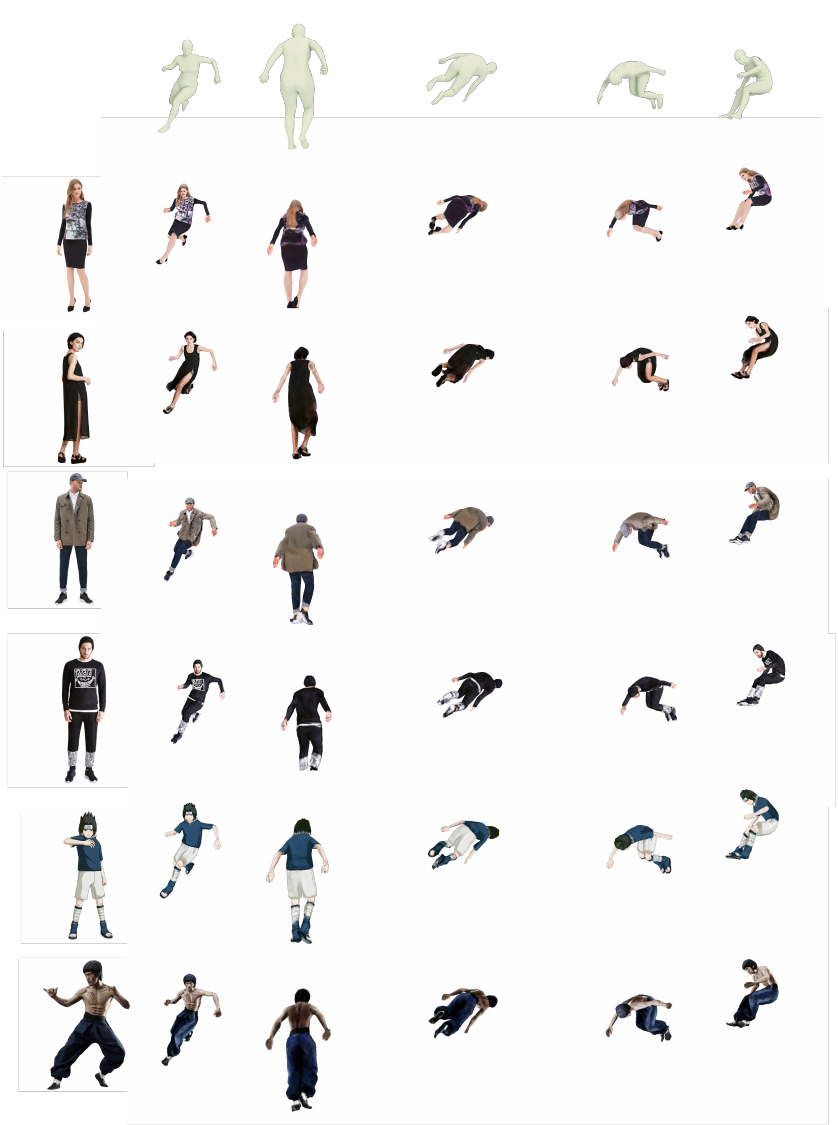}
    \\
    \makebox[0.08\textwidth]{\footnotesize Reference}
    \makebox[0.8\textwidth]{\footnotesize Animation results} 
   \\ 
    \caption{Visual results of human animation results~(Part III) from any input. Best viewed with zoom-in.} 
    \label{fig:animation_v3}
\end{figure*}

\begin{figure*}[tb] \centering
    \includegraphics[width=\textwidth]{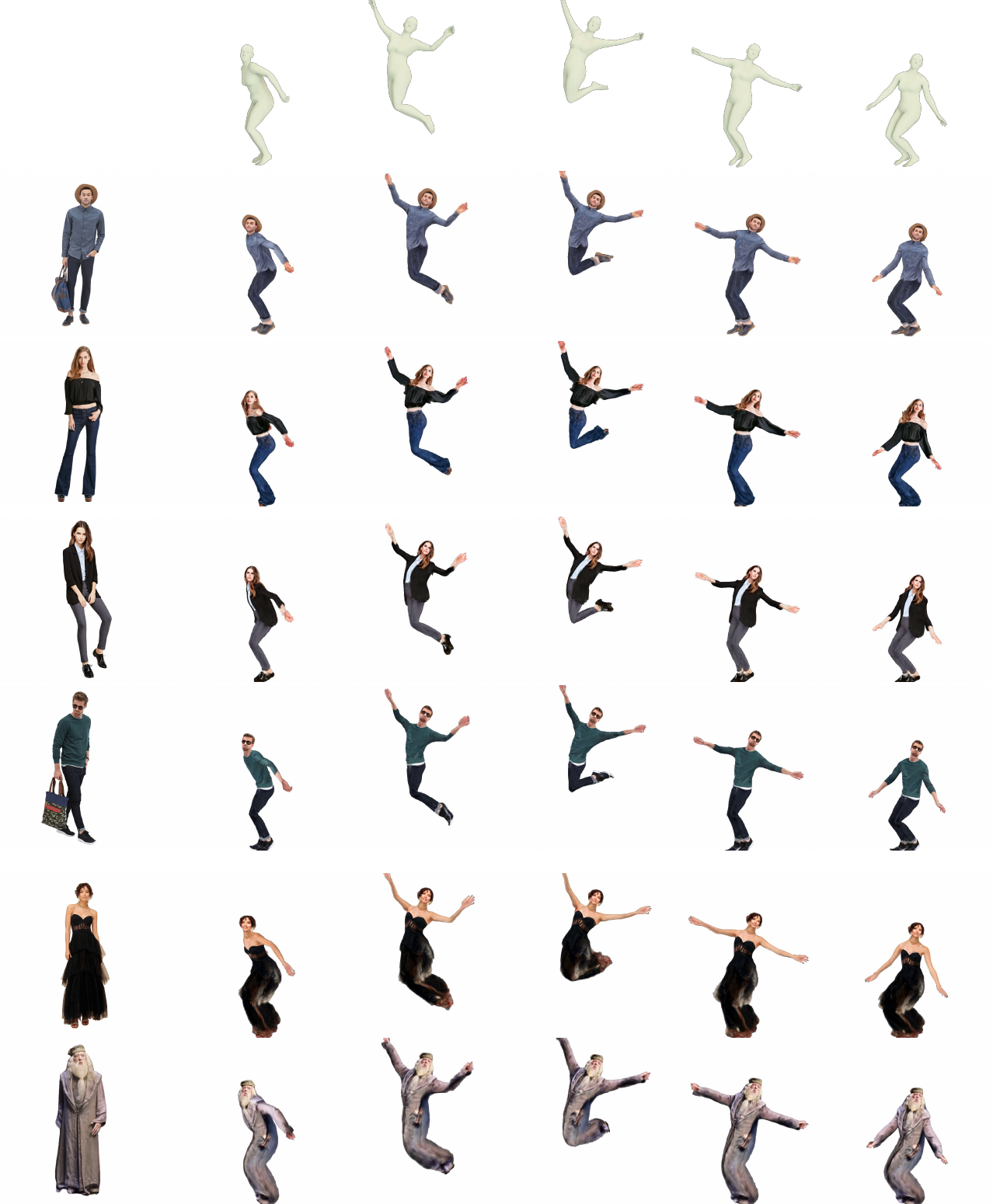}
    \\
    \makebox[0.08\textwidth]{\footnotesize Reference}
    \makebox[0.8\textwidth]{\footnotesize Animation results} 
   \\ 
    \caption{Visual results of human animation results~(Part IV) from any input. Best viewed with zoom-in.} 
    \label{fig:animation_v4}
\end{figure*}

\subsection{Reconstruction and Animation from Any Input}
\begin{figure*}[tb] \centering
    \includegraphics[width=\textwidth]{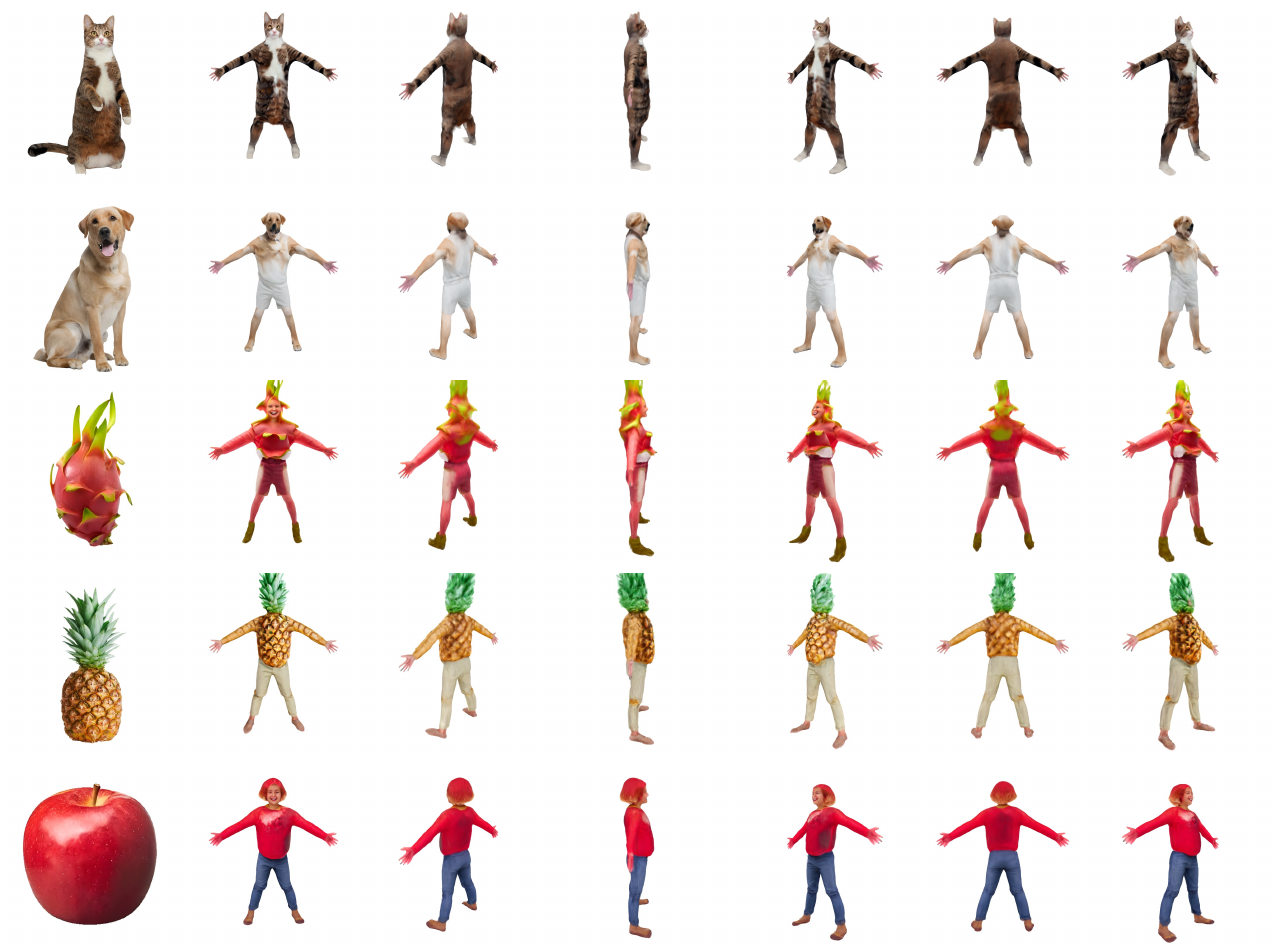}
    \\
    \makebox[0.1\textwidth]{\footnotesize Reference}
    \makebox[0.8\textwidth]{\footnotesize Multi-view Reconstruction} 
   \\ 
    \caption{Visual results of canonical shape reconstruction from ``Any Input''. Best viewed with zoom-in.} 
    \label{fig:vis_a_anyinput}
\end{figure*}

\begin{figure*}[tb] \centering
    \includegraphics[width=\textwidth]{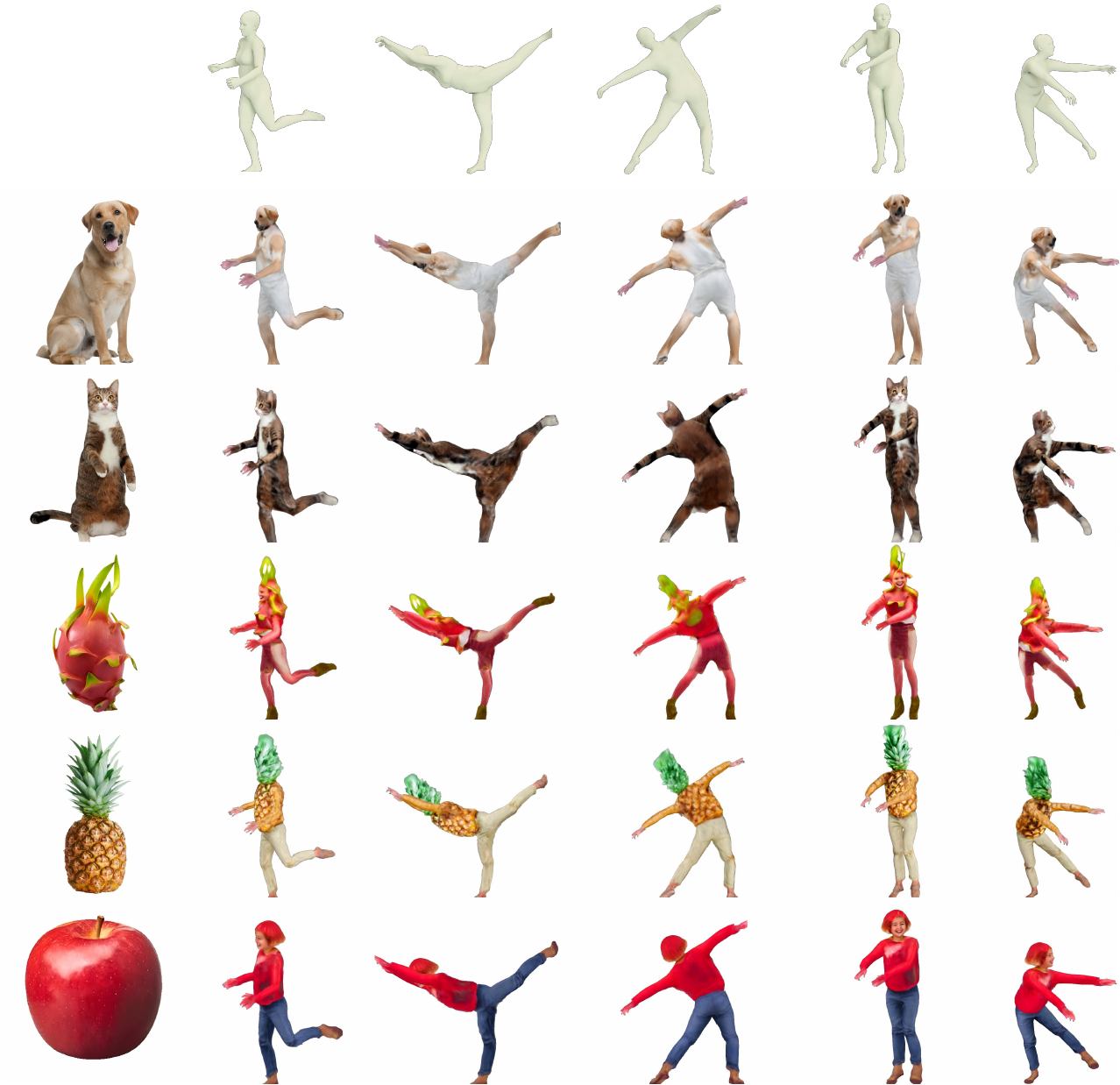}
    \\
    \makebox[0.1\textwidth]{\footnotesize Reference}
    \makebox[0.8\textwidth]{\footnotesize Multi-view Reconstruction} 
   \\ 
    \caption{Visual results of canonical shape reconstruction from ``Any Input''. Best viewed with zoom-in.} 
    \label{fig:vis_animated_animals}
\end{figure*}

\Fref{fig:vis_a_anyinput} and \fref{fig:vis_animated_animals} illustrate reconstructions and animation results from a diverse set of images collected from the internet. Notably, the reference image is a non-human image input, demonstrating the model's still maintain original diffusion model's generalizability.

\subsection{Canonical Shape Reconstruction} 
\begin{figure*}[tb] \centering
    \includegraphics[width=\textwidth]{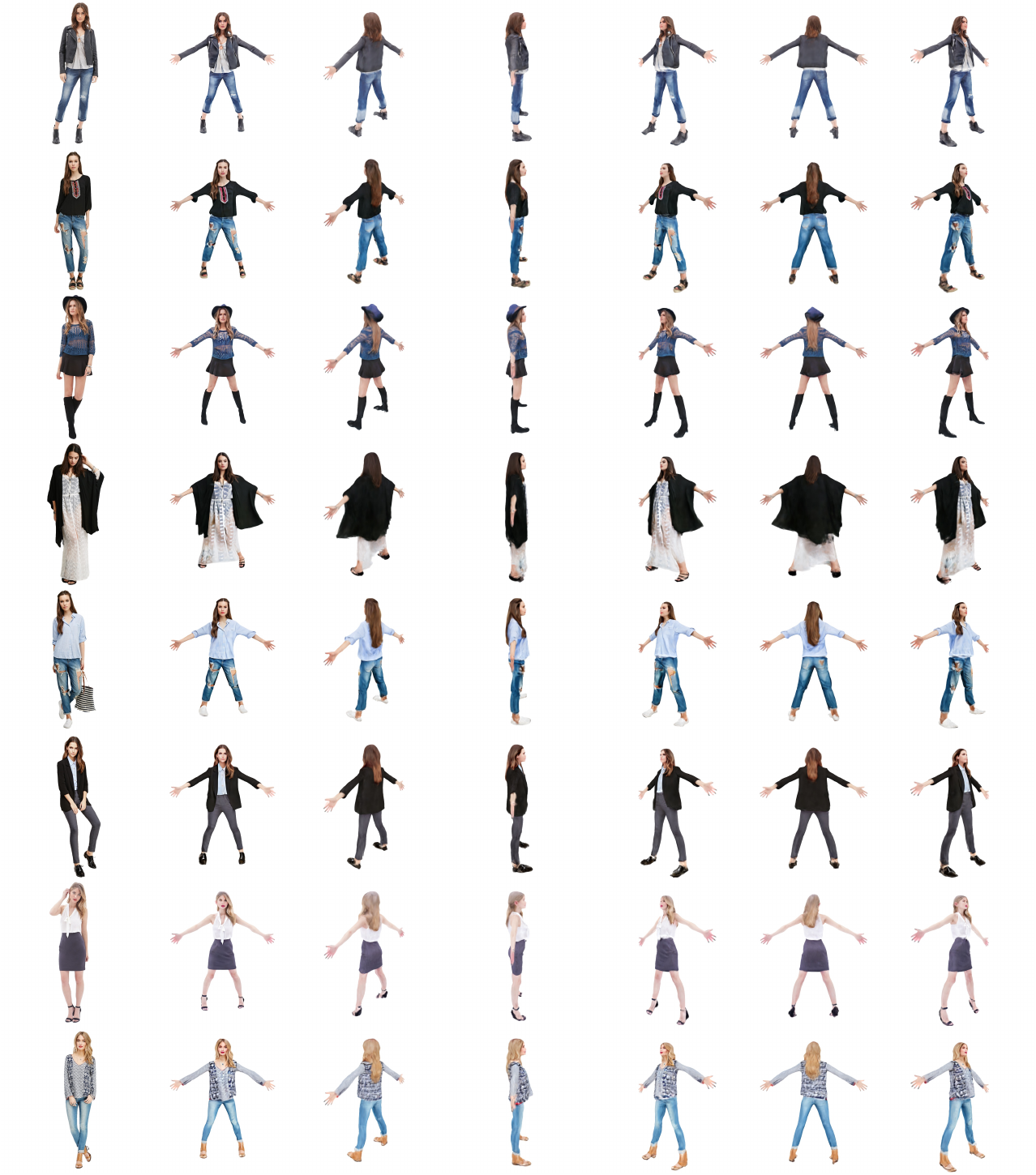}
    \\
    \makebox[0.1\textwidth]{\footnotesize Reference }
    \makebox[0.8\textwidth]{\footnotesize Multi-view Reconstruction} 
   \\ 
    \caption{Visual results of canonical shape reconstruction~(Part I). Best viewed with zoom-in.} 
    \label{fig:vis_a_1}
\end{figure*}

\begin{figure*}[tb] \centering
    \includegraphics[width=\textwidth]{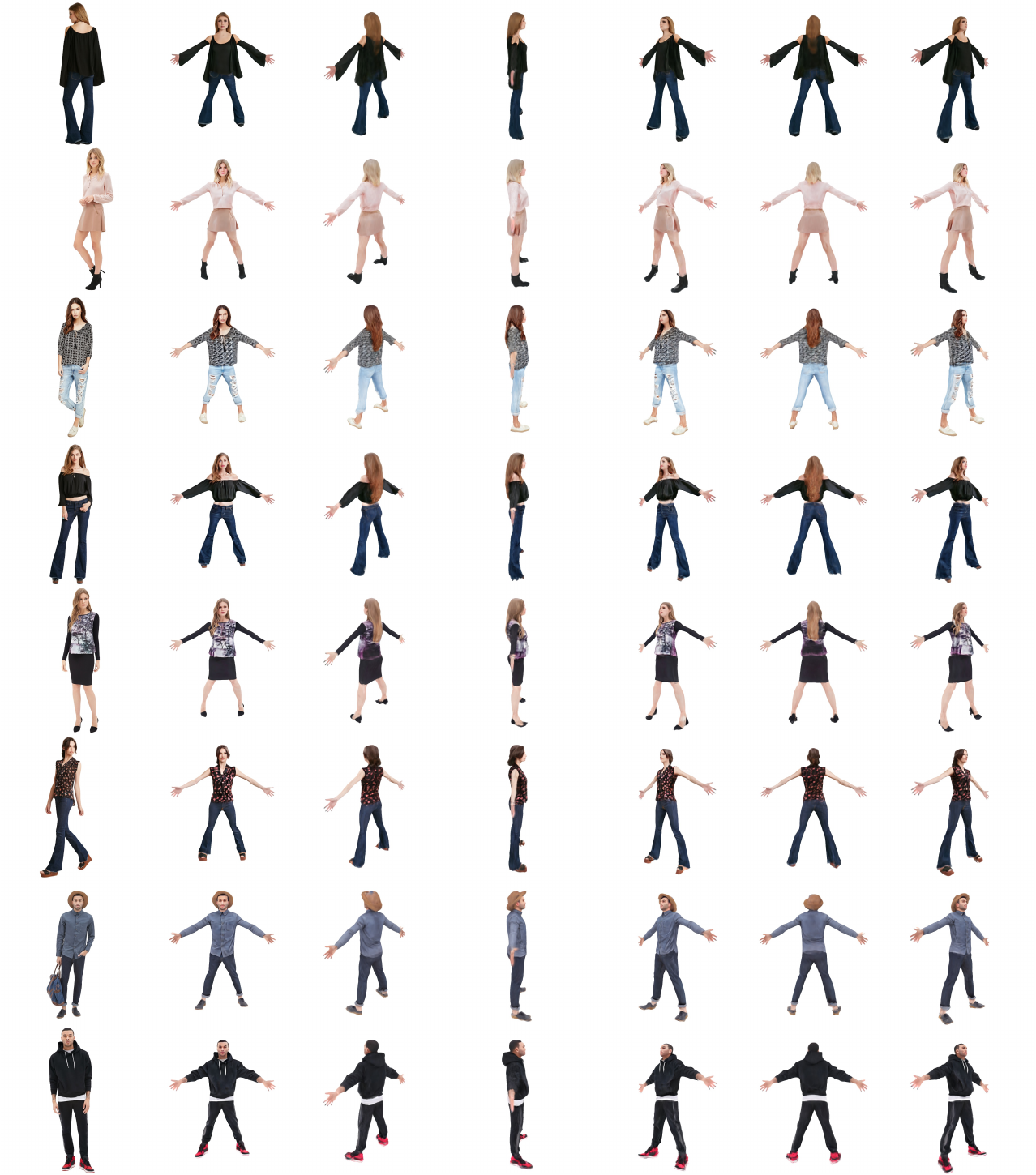}
    \\
    \makebox[0.1\textwidth]{\footnotesize Reference }
    \makebox[0.8\textwidth]{\footnotesize Multi-view Reconstruction} 
   \\ 
    \caption{Visual results of canonical shape reconstruction~(Part II). Best viewed with zoom-in.} 
    \label{fig:vis_a_2}
\end{figure*}

\begin{figure*}[tb] \centering
    \includegraphics[width=\textwidth]{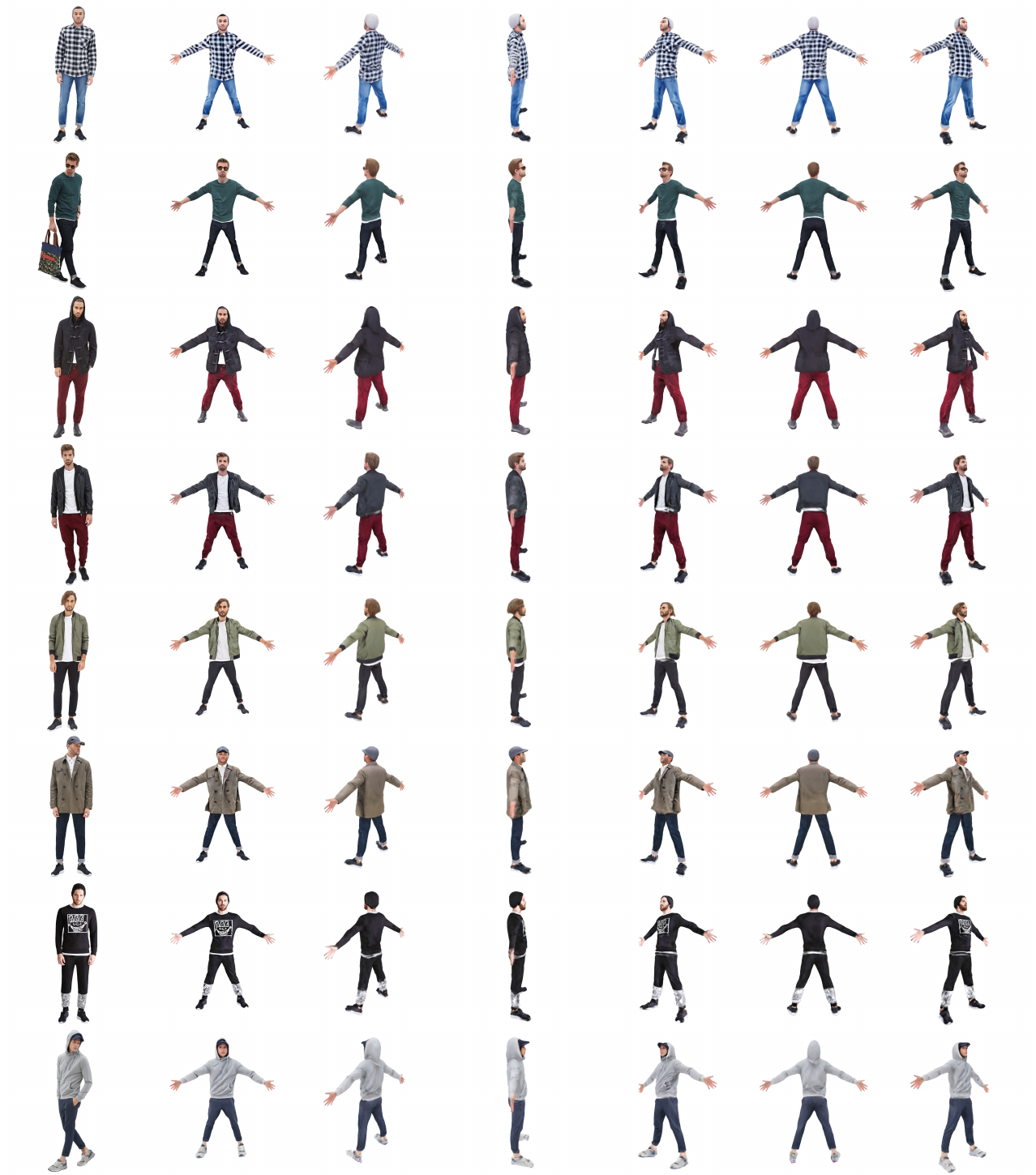}
    \\
    \makebox[0.1\textwidth]{\footnotesize Reference }
    \makebox[0.8\textwidth]{\footnotesize Multi-view Reconstruction} 
   \\ 
    \caption{Visual results of canonical shape reconstruction~(Part III). Best viewed with zoom-in.} 
    \label{fig:vis_a_3}
\end{figure*}

\begin{figure*}[tb] \centering
    \includegraphics[width=\textwidth]{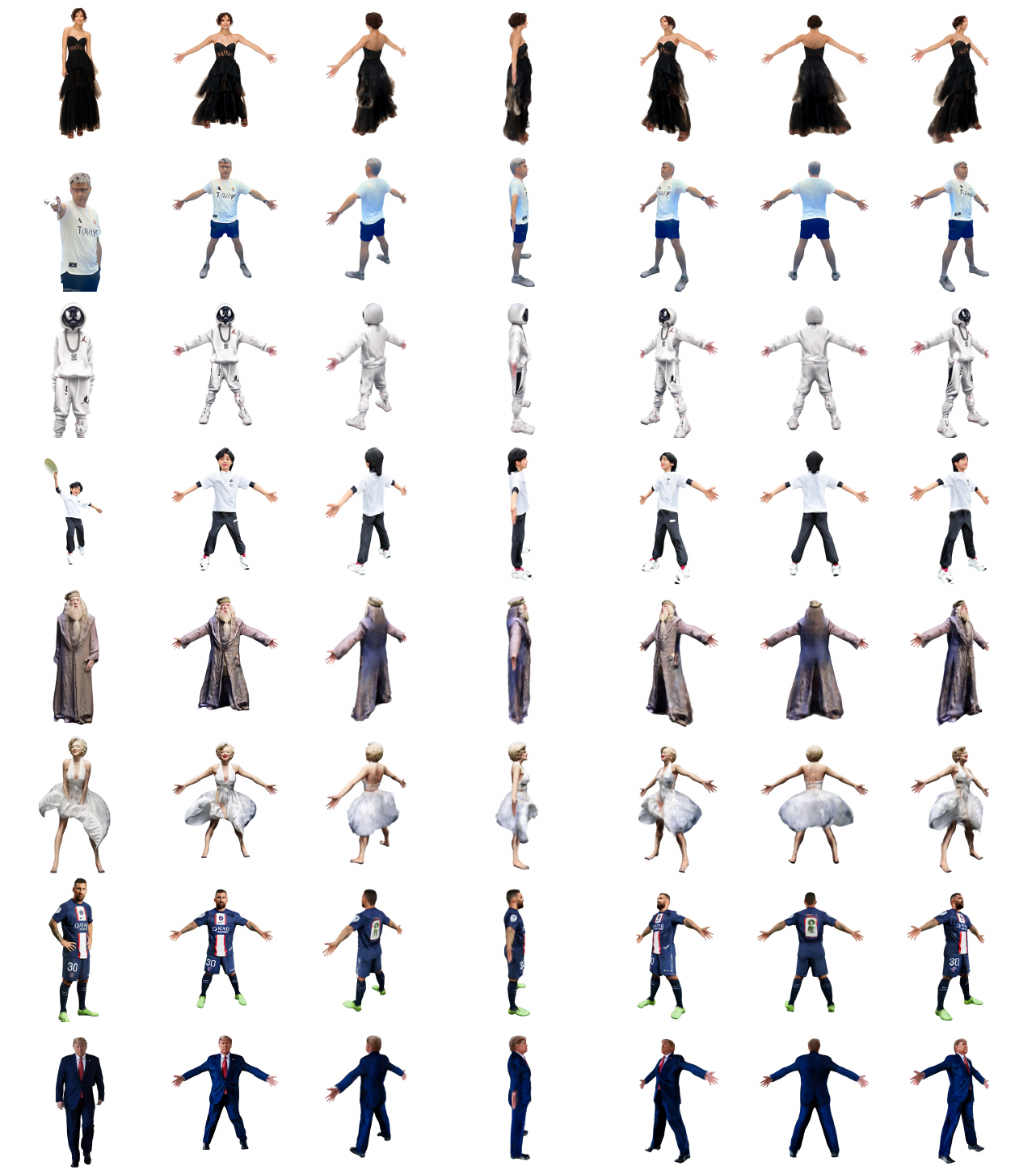}
    \\
    \makebox[0.1\textwidth]{\footnotesize Reference }
    \makebox[0.8\textwidth]{\footnotesize Multi-view Reconstruction} 
   \\ 
    \caption{Visual results of canonical shape reconstruction~(Part IV). Best viewed with zoom-in.} 
    \label{fig:vis_a_4}
\end{figure*}

\begin{figure*}[tb] \centering
    \includegraphics[width=\textwidth]{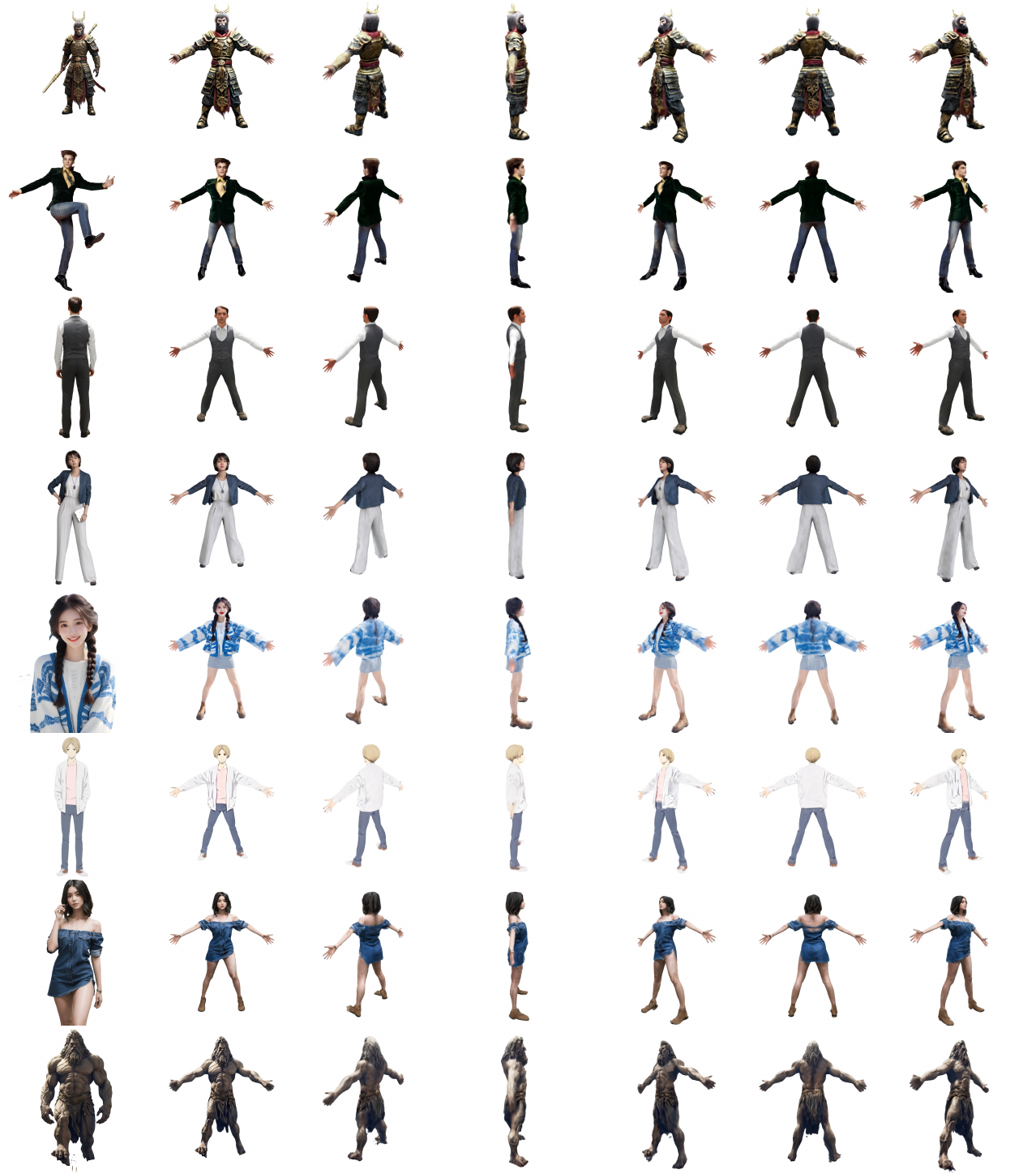}
    \\
    \makebox[0.1\textwidth]{\footnotesize Reference }
    \makebox[0.8\textwidth]{\footnotesize Multi-view Reconstruction} 
   \\ 
    \caption{Visual results of canonical shape reconstruction~(Part V). Best viewed with zoom-in.} 
    \label{fig:vis_a_5}
\end{figure*}

\begin{figure*}[tb] \centering
    \includegraphics[width=\textwidth]{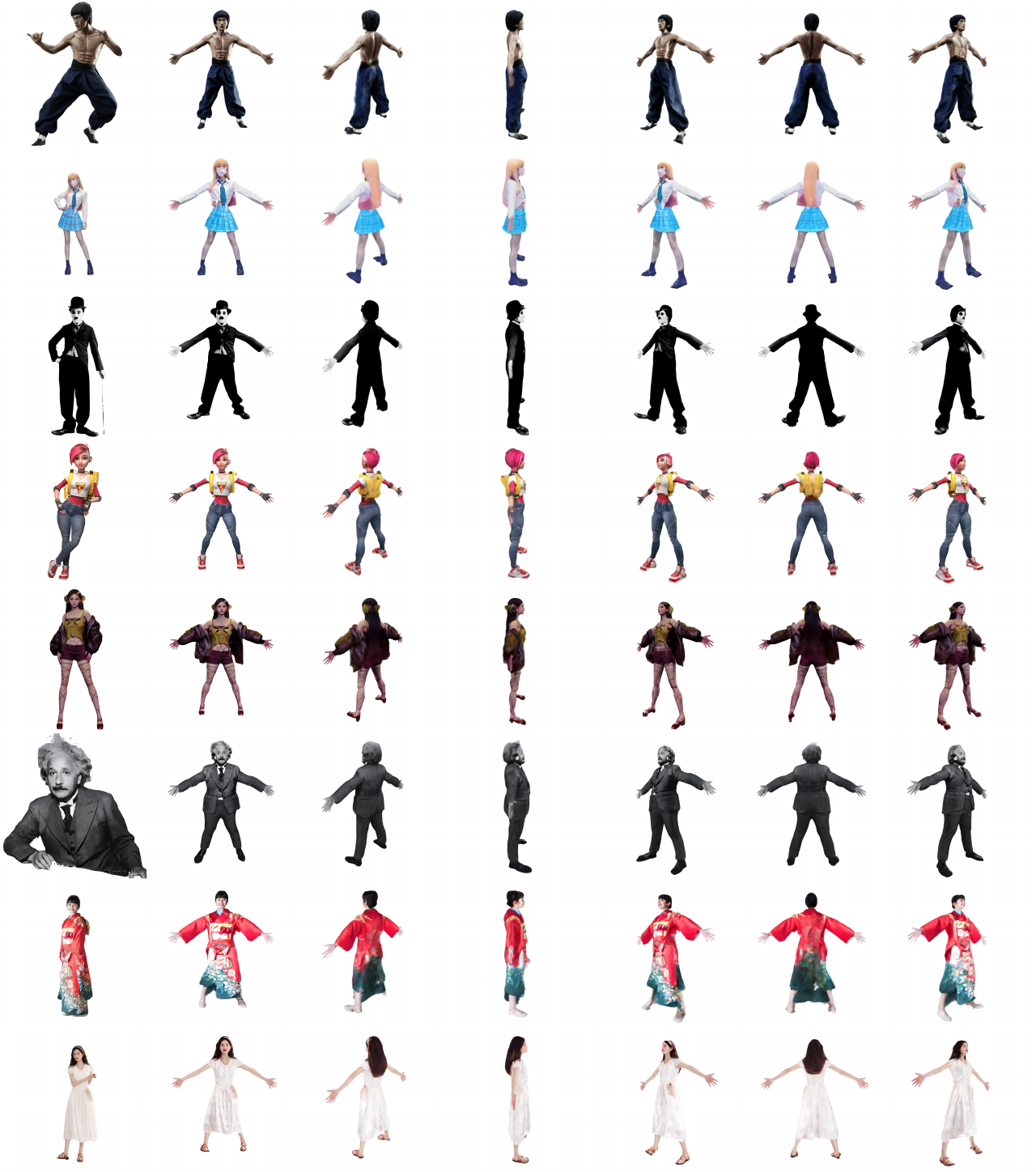}
    \\
    \makebox[0.1\textwidth]{\footnotesize Reference }
    \makebox[0.8\textwidth]{\footnotesize Multi-view Reconstruction} 
   \\ 
    \caption{Visual results of canonical shape reconstruction~(Part VI). Best viewed with zoom-in.} 
    \label{fig:vis_a_6}
\end{figure*}

To further validate the effectiveness of the proposed method, we provide additional results for canonical shape reconstruction from single images. 
\Fref{fig:vis_a_1}--\fref{fig:vis_a_3} present reconstruction results on the DeepFashion dataset, showcasing accurate recovery of canonical shapes from fashion images. 
Meanwhile, \Fref{fig:vis_a_4}--\fref{fig:vis_a_6} illustrate reconstructions from a diverse set of images collected from the internet, demonstrating the model's adaptability to various image sources and styles.

